\documentclass{article}

\PassOptionsToPackage{numbers, compress}{natbib}
\usepackage[preprint]{neurips_2026}

\usepackage{hyperref}
\usepackage{url}
\usepackage{nicefrac}
\usepackage{xcolor}
\usepackage{wrapfig}
\usepackage[T1]{fontenc}
\usepackage{arydshln}

\usepackage[utf8]{inputenc}

\usepackage{microtype}

\usepackage{inconsolata}
\usepackage{graphicx}
\usepackage{enumitem}

\usepackage{amsmath}
\usepackage{booktabs} 
\usepackage{multirow}
\usepackage[table]{xcolor}
\usepackage{amsfonts}
\usepackage{svg}

\usepackage{subcaption}

\usepackage{amsthm}
\theoremstyle{definition}

\makeatletter
\renewcommand{\@noticestring}{%
  Accepted for publication at NeurIPS 2026.%
}
\makeatother

\title{Foresight-over-Graph: Reasoning Beyond Local Horizons for Knowledge Base Question Answering}

\author{
\textbf{Yang Hong}\textsuperscript{1},
\textbf{Yajun Yang}\textsuperscript{1}\thanks{Corresponding author.},
\textbf{Xin Wang}\textsuperscript{1},
\textbf{Liping Jing}\textsuperscript{2},
\textbf{Qinghua Hu}\textsuperscript{1}
\\
\textsuperscript{1}Tianjin University, Tianjin, China \\
\textsuperscript{2}Beijing Jiaotong University, Beijing, China \\
\texttt{\{yanghong, yjyang, wangx, huqinghua\}@tju.edu.cn},\\
\texttt{lpjing@bjtu.edu.cn}
}

\begin{document}
\maketitle
\begin{abstract}
Large language models (LLMs) have demonstrated strong capabilities in question answering, yet they still frequently suffer from hallucinations on knowledge-intensive tasks. Knowledge graphs (KGs) provide LLMs with structured, interpretable, and updatable factual grounding, making them a promising external knowledge source for reliable reasoning. However, existing LLM-guided graph reasoning methods typically rely on hop-wise greedy or beam-style pruning during evidence retrieval. Such local decision processes are inherently myopic: evidence that appears weak near the source may become crucial only after deeper graph context is explored, causing answer-critical branches to be discarded prematurely and making the reasoning chain difficult to recover. To address this limitation, we propose Foresight-over-Graph (FoG), a foresight-aware evidence retrieval framework for knowledge base question answering (KBQA). FoG iteratively constructs a question-relevant evidence subgraph and uses far-to-near feedback to guide path exploration, and maintains a compact memory subgraph to support continued exploration. Extensive experiments on widely used KBQA benchmarks demonstrate that FoG achieves state-of-the-art performance, with a particularly large improvement of 16.58\% in Hit on CWQ, while also reducing LLM calls and token usage. Our code is available at \url{https://github.com/yhong7/FoG}
.
\end{abstract}
\section{Introduction}

Large language models (LLMs) have recently demonstrated strong capabilities in question answering, yet they often hallucinate when questions require reliable professional or domain-specific knowledge \cite{guan2024mitigating, mallen2023trust, kandpal2023large}. A straightforward way to inject such knowledge is to continually fine-tune LLMs, but this is typically expensive and inefficient, and still struggles to keep models synchronized with evolving real-world facts \cite{wang2024knowledge, ni2024forgetting}. In contrast, knowledge graphs (KGs) provide structured, updatable, and interpretable factual knowledge, offering an effective way to ground LLM-based QA and reduce hallucinations \cite{ji2021survey, zhang2024knowgpt, pan2024unifying, pusch2024combining, lavrinovics2025knowledge}.

Recent progress has led to a spectrum of KG-integrated LLM approaches 
\cite{jiang2023structgpt, zhang2024knowgpt, wang2024kgp, he2024gretriever}.
The core idea of these methods is to retrieve question-relevant entities, relations, or subgraphs from the knowledge graph, and then feed the retrieved results into the LLM to generate the final answer 
\cite{he2024gretriever,  zhang2024knowgpt}.
To better leverage the capabilities of LLMs, recent works propose LLM-guided graph agents that perform search over knowledge graphs hop by hop to enable retrieval-augmented generation 
\cite{chen2024pog, liang2025fasttog}.
A well-designed LLM-guided agent iteratively chooses which part of the graph to explore, gathering relevant evidence to generate the final answer 
\cite{guo2024knowledgenavigator, chen2024pog, liang2025fasttog}.



\begin{wrapfigure}{r}{0.48\linewidth}
    \centering

    \begin{subfigure}[t]{\linewidth}
        \centering
        \includegraphics[width=\linewidth]{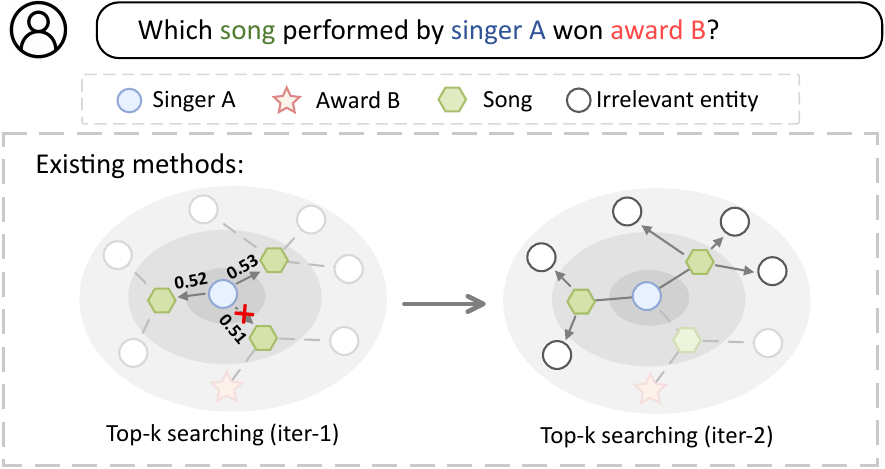}
        \caption{Existing methods: local decisions can drop crucial triples early, causing later steps to drift into irrelevant regions.}
        \label{fig:example-a}
    \end{subfigure}

    \vspace{4pt}

    \begin{subfigure}[t]{\linewidth}
        \centering
        \includegraphics[width=\linewidth]{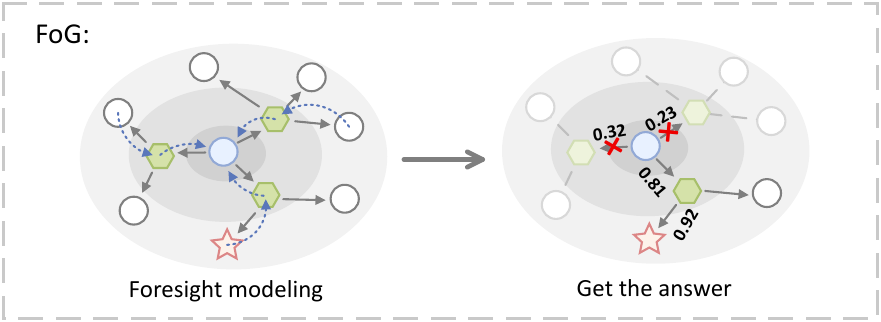}
        \caption{FoG: injects foresight from the expanded frontier to refine retrieval decisions.}
        \label{fig:example-b}
    \end{subfigure}

    \caption{Limitations of short-sighted hop-by-hop KG traversal and the intuition behind FoG.}
    \label{fig:example}
\end{wrapfigure}

However, KGs are typically large-scale and a substantial fraction of nodes are associated with a large number of relations\cite{xu-etal-2024-generate,shen-etal-2025-reasoning}. This makes it intractable to expand all outgoing relations at each hop, because deep multi-hop reasoning leads to an explosion of candidate nodes. Most previous works adopt greedy or beam-style traversal and perform local pruning, keeping only an appropriately sized set of candidate relations (e.g., top-$k$) at each search step according to the relevance between the question and candidate relations \cite{thinkongraph, luo2023rog}. Although these methods demonstrate strong performance, greedy-based and beam-search-based per-hop pruning still struggles with complex questions that require deep exploration over KGs. The reason is that, in the early stages of search, many relations cannot be distinguished with foresight as to whether they lie on the answer path, which may cause truly answer-relevant relations to be pruned away. Figure~\ref{fig:example-a} shows a toy example, where the question is "Which song performed by singer A won award B?". 
At the first hop, all the neighbors of the source entity (i.e., singer A) are songs performed by singer A with no significant difference in relevance to the question.
Existing methods cannot identify which branch leads to the correct answer at this stage, since they have no access to distant entities or evidence (e.g., Award B). As a result, correct triples may be pruned away.
\textbf{Once the critical path is mistakenly pruned, subsequent exploration cannot recover the missing evidence chain, and the search often drifts into irrelevant neighborhoods.}
A real case study illustrating this failure mode is provided in Appendix~\ref{app:case-study}.

To address this limitation, we propose a novel \textbf{Foresight-over-Graph (FoG)} model that identifies which relations are more important for answering the question, thereby avoiding incorrect pruning, as shown in Figure~\ref{fig:example-b}. Given a Q\&A task, FoG first extracts source entities from the question and then iteratively expands subgraph from these entities until the answer is found. In each iteration, triples that are irrelevant to the question are discarded, yielding an expanded subgraph with high recall and low noise. We further design a far-to-near message feedback mechanism that propagates question-relevant information from distant entities to entities closer to the source entities, and utilize this information to re-score and prune nearby triples. Finally, the LLM incorporates the graph structure refined by question relevance into its memory and generates new sub-questions to guide the next search iteration. Our main contributions are summarized as follows:






\begin{itemize}
\item \textbf{Bottleneck Identification.} We identify a critical bottleneck in existing graph-guided LLM agents: hop-wise greedy/beam pruning is \textbf{myopic}, lacking foresight into distant evidence and leading to irreversible loss of answer-critical triples.
    
    \item \textbf{Proposed Method.} We propose \textbf{Foresight-over-Graph (FoG)}, which counteracts this short-sightedness with two designs: a far-to-near message feedback mechanism that guides early candidates using distant signals, and a memory subgraph manager that preserves high-confidence evidence across iterations.
    
    \item \textbf{Empirical Analysis.} On WebQSP and CWQ, FoG achieves state-of-the-art results, gaining 16.58\% in Hit on CWQ while reducing LLM calls and token usage against existing LLM-guided graph agents.
\end{itemize}

\section{Preliminaries}
\label{sec:preliminaries}

\paragraph{Structured Knowledge Base.}
We consider a structured knowledge base represented as a multi-relational directed graph
$\mathcal{G}=\{\langle s,r,o\rangle \mid s,o\in\mathcal{E}, r\in\mathcal{R}\}$,
where $\mathcal{E}$ denotes the entity set and $\mathcal{R}$ denotes the relation set.
Each triple $\langle s,r,o\rangle$ states that relation $r$ holds between subject entity $s$ and object entity $o$.
This graph representation is widely used in KBQA benchmarks and provides an explicit, interpretable substrate for multi-hop evidence retrieval.

\paragraph{Problem Definition.}
Given a complex natural-language question $q_{\mathrm{ori}}$, a structured knowledge base $\mathcal{G}$,
and a set of source entities $\mathcal{E}_s \subseteq \mathcal{E}$ identified from the question,
the goal of KBQA is to predict an answer set $\mathcal{A}\subseteq\mathcal{E}$.
Rather than producing answers from parametric knowledge alone, a faithful KBQA system should retrieve
a compact evidence subgraph $\mathcal{G}_{\mathrm{ans}}\subseteq\mathcal{G}$ originating from $\mathcal{E}_s$
that semantically and factually supports the final prediction.

\section{Method}
\label{sec:method}

\subsection{Foresight Subgraph Expansion}
\label{subsec:expansion}

The initial stage of FoG aims to extract a high-recall, low-noise expanded subgraph $\mathcal{G}^{\mathrm{xpd}}$ from the knowledge graph. By filtering incident edges via a foresight triple scorer, we ensure that subsequent message passing operates on a compact region of informative triples, thereby avoiding the explosion of candidate nodes during deep multi-hop reasoning.

\noindent\textbf{Question-Conditioned Triple Encoder.}
We leverage a shared pre-trained embedding model to represent KG symbols and text queries. For each ordered triple instance $t=\langle s, r, o \rangle$, we augment the base embeddings with learnable role-specific biases to form a sequence $\mathbf{u}_t = [\mathbf{e}_s+\mathbf{p}_s;\ \mathbf{e}_r+\mathbf{p}_r;\ \mathbf{e}_o+\mathbf{p}_o] \in \mathbb{R}^{3 \times d}$. Rather than detailing standard Transformer layers, we define the encoding process as a dual-attention mechanism. We first apply a self-attention block to model intra-triple factual interactions:
\begin{equation}
\mathbf{u}^{\mathrm{self}}_t = \mathrm{SelfAttn}(\mathbf{u}_t).
\end{equation}
Subsequently, a cross-attention block is employed to inject query-specific semantics into the representation:
\begin{equation}
\mathbf{z}_t = \mathrm{CrossAttn}(\text{query}=\mathbf{q}_{\mathrm{ori}}, \text{key/value}=\mathbf{u}^{\mathrm{self}}_t).
\end{equation}
The resulting question-guided embedding $\mathbf{z}_t$ serves as a semantically grounded representation of the triple and remains fixed during iterative reasoning rounds.

\noindent\textbf{Foresight Triple Scorer.}
The original question $\mathbf{q}_{\mathrm{ori}}$ is often too coarse-grained for multi-hop search. We therefore prompt an LLM to decompose it into a set of fine-grained sub-questions $\mathcal{Q}_{\mathrm{sub}}$. Each $q \in \mathcal{Q}_{\mathrm{sub}}$ represents a concrete sub-goal necessary for solving the original query.

To prioritize triples that lead to the true answer path, we define the structural context of triple $t$ as $\mathbf{c}_t = [\mathbf{z}_t; \mathbf{h}_s; \mathbf{h}_o]$, where $\mathbf{h}_v$ denotes the foresight hidden state of entity $v$ and is initialized as zero vector. We then employ an attention-pooling mechanism to dynamically aggregate the relevant sub-questions:
\begin{equation}
\bar{\mathbf{q}}_t = \sum_{q \in \mathcal{Q}_{\mathrm{sub}}} \alpha_{t,q} \mathbf{q}, \quad \text{where} \ \alpha_{t,q} = \mathrm{Softmax}(\mathcal{F}_{\mathrm{attn}}(\mathbf{c}_t, \mathbf{q})).
\end{equation}
This query-aware pooled representation is then combined with the structural context to yield the unified foresight score:
\begin{equation}
s_t = \mathrm{MLP}_s([\mathbf{c}_t; \bar{\mathbf{q}}_t]).
\end{equation}
This formulation allows the scorer to favor triples that satisfy both the local structural constraints and the global reasoning goals reflected in the sub-questions.

\noindent\textbf{Foresight Subgraph Construction.}
Starting from the seed entities $\mathcal{E}_s$, we perform a layered Breadth-First Search (BFS) expansion up to $L$ hops. At each frontier layer, we evaluate the incident triples and retain those exceeding an expansion threshold $\tau_e$:
\begin{equation}
\mathcal{G}_{\mathrm{layer}} = \big\{ t \in \mathcal{N}(v) \mid s_t > \tau_e \big\}.
\end{equation}
To facilitate subsequent message feedback, the retrieved region is mapped into a directed subgraph $\mathcal{G}^{\mathrm{xpd}}$. Specifically, each triple $t=\langle s, r, o \rangle$ is oriented inward toward the seeds based on the endpoints' shortest-path distances. This directionality enforces far-to-near propagation for propagating lookahead signals from distant evidence back to source-side candidates.

\subsection{Foresight Feedback Propagation}
\label{subsec:foresight-msg}

We introduce a foresight feedback propagation module to inject signals from the compressed future search space into early decisions, updating node-level hidden states and refining edge scores.

\begin{figure*}[t]
    \centering
    \hspace*{2mm}
    \includegraphics[width=1.0\textwidth]{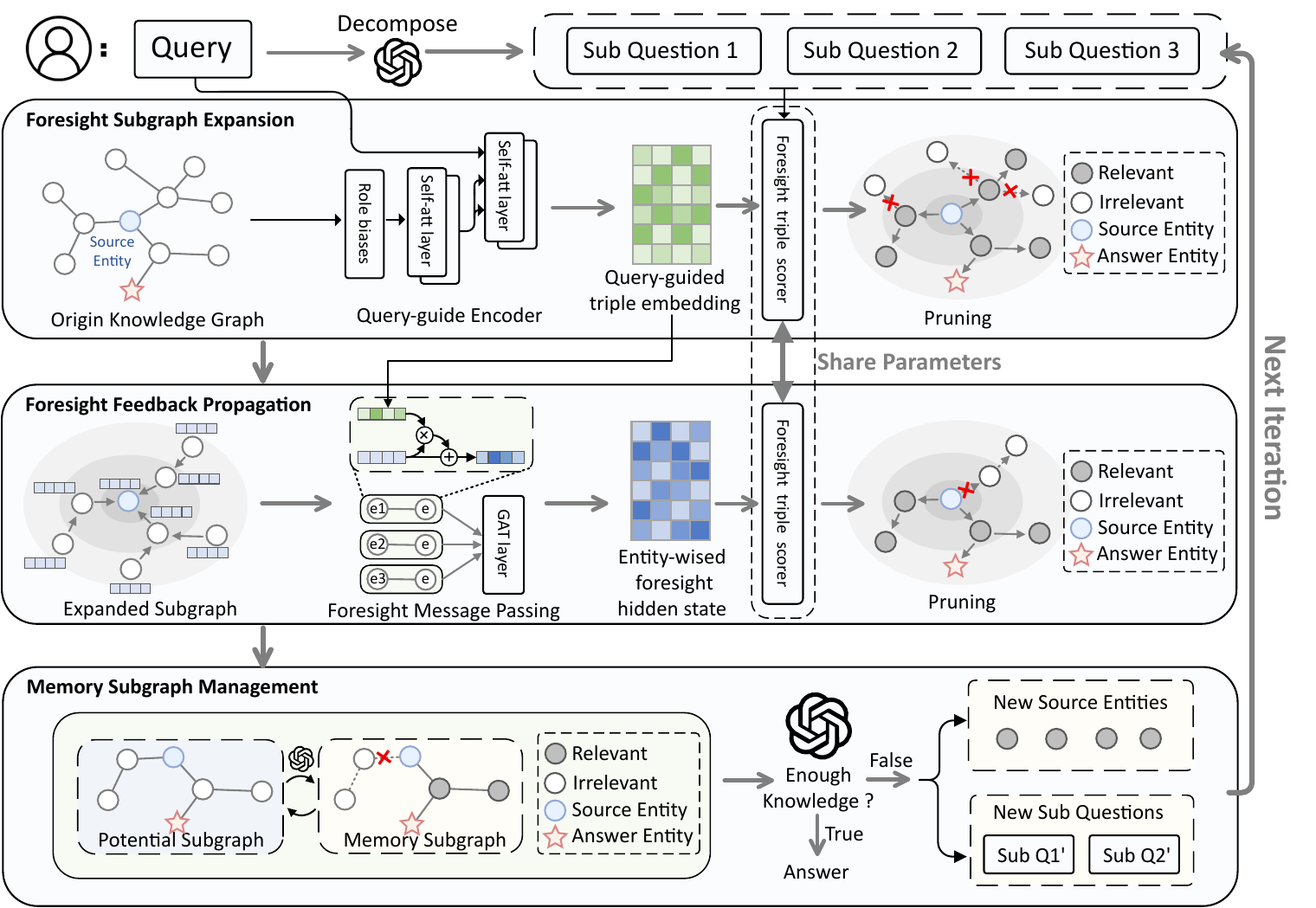}
    \caption{The overview of our proposed FoG.}
    \label{fig:framework}
\end{figure*}

\noindent\textbf{Foresight Feedback Propagation.}
To propagate foresight information from the outer expansion frontier back to the source, we operate on the directed subgraph $\mathcal{G}^{\mathrm{xpd}}$. Nodes are updated in decreasing order of their distance $d(\cdot)$, ensuring a strict far-to-near information flow. 

For each inward-directed triple $t=\langle s, r, o \rangle$, we compute an edge-wise future message by fusing the source entity's hidden state $\mathbf{h}_s$ with the question-guided triple embedding $\mathbf{z}_t$:
\begin{equation}
\mathbf{m}_t = \mathbf{h}_s + \mathrm{MLP}_m([\mathbf{h}_s; \mathbf{z}_t]).
\end{equation}
Next, at each target entity $o$, we gather incoming messages from its neighborhood $\mathcal{G}^{\text{in}}(o)$. We compute compatibility scores between each message and the target's current state $\mathbf{h}_o$:
\begin{equation}
\alpha_{t,q} =
\frac{\exp(F_{attn}(c_t,q))}
{\sum_{q' \in Q_{sub}}\exp(F_{attn}(c_t,q'))}
\end{equation}
The target entity then updates its representation by assimilating the attention-weighted messages through a non-linear activation $\phi$:
\begin{equation}
\mathbf{h}_o \leftarrow \mathbf{h}_o + \phi\bigg(\sum_{t\in\mathcal{G}^{\text{in}}(o)} \alpha_t \mathbf{m}_t\bigg).
\end{equation}
After all entities are updated layer by layer, we reuse the foresight scorer to re-evaluate all edges. These refined scores $\{s_t\}$ are subsequently used to extract the high-confidence potential subgraph for the LLM's memory management.

\noindent\textbf{Training.}
\label{subsec:training}
The foresight scorer is trained as a binary classifier over candidate triples.
For each training question $q_{\mathrm{tr}}$, let $\mathcal{G}^{+}(q_{\mathrm{tr}})$ be the set of triples on the ground-truth reasoning path.
The label of a triple $t$ is $y_t = \mathbb{I}[t \in \mathcal{G}^{+}(q_{\mathrm{tr}})]$.
We minimize the binary cross-entropy loss over all candidates in the training graph $\mathcal{G}^{\mathrm{tr}}$:
\begin{equation}
\mathcal{L}_{\mathrm{tr}} = -\sum_{t\in\mathcal{G}^{\mathrm{tr}}} \big( y_t \log \sigma(s_t) + (1-y_t) \log (1-\sigma(s_t)) \big).
\end{equation}
Details of candidate sampling and training setup are in Appendix~\ref{sec:appendix-training}.

\subsection{Memory Subgraph Management}
\label{subsec:memory}

In this subsection, we detail how a persistent memory subgraph, orchestrated by the LLM, accumulates high-confidence evidence. This mechanism prevents short-sighted pruning errors and directs subsequent retrieval iterations toward unexplored or uncertain regions of the knowledge graph.

\noindent\textbf{Constructing the Potential Subgraph.}
Instead of overwhelming the LLM with the entire expanded region, we isolate a compact, high-confidence candidate set. Using the refined foresight scores $\{s_t\}$ obtained after feedback propagation, we extract a potential subgraph by applying a potential threshold $\tau_{\mathrm{p}}$ to retain only the most promising candidates:
\begin{equation}
\mathcal{G}^{\mathrm{pot}} = \big\{ t \in \mathcal{G}^{\mathrm{xpd}} \mid s_t > \tau_{\mathrm{p}} \big\}.
\end{equation}
Only this highly relevant graph context $\mathcal{G}^{\mathrm{pot}}$ is exposed to the LLM for high-level reasoning, significantly reducing the context window burden and filtering out structural noise.

\noindent\textbf{LLM-Guided Memory Update.}
We maintain a persistent memory subgraph $\mathcal{G}^{\mathrm{mem}} \subseteq \mathcal{G}$ (initialized as $\emptyset$) to selectively archive triples deemed globally useful for answering the query. In each iteration, we formulate a structured prompt containing the original question $q_{\mathrm{ori}}$, the historical memory $\mathcal{G}^{\mathrm{mem}}$, and the newly discovered potential subgraph $\mathcal{G}^{\mathrm{pot}}$. 

We formalize the LLM's reasoning process as a generation function that simultaneously outputs three structured components for memory update, tentative answering, and continued exploration:
\begin{equation}
(\mathcal{G}^{\mathrm{mem}\prime}, \hat{A}, \mathcal{Q}_{\mathrm{sub}}^{\prime}) = \mathrm{LLM}\big(q_{\mathrm{ori}}, \mathcal{G}^{\mathrm{mem}}, \mathcal{G}^{\mathrm{pot}}\big).
\end{equation}
Here, $\mathcal{G}^{\mathrm{mem}\prime}$ represents the updated memory retaining crucial evidence, $\hat{A}$ denotes the tentative answer set, and $\mathcal{Q}_{\mathrm{sub}}^{\prime}$ contains newly generated sub-questions designed to drive the next search phase. Following this interaction, we strictly overwrite the historical state:
\begin{equation}
\mathcal{G}^{\mathrm{mem}} \leftarrow \mathcal{G}^{\mathrm{mem}\prime}.
\end{equation}
Detailed prompt templates and output JSON schemas are provided in Appendix~\ref{sec:appendix-llm}.

\noindent\textbf{Exploration Efficiency and Fault Tolerance.}
By coupling iterative subgraph re-expansion with persistent memory maintenance, FoG naturally avoids irreversible early pruning.
The memory subgraph $\mathcal{G}^{\mathrm{mem}}$ continuously archives high-confidence evidence across rounds, while newly generated sub-questions $\mathcal{Q}_{\mathrm{sub}}^{\prime}$ drive targeted re-exploration toward unresolved sub-goals.
Consequently, a triple that scores below $\tau_{\mathrm{p}}$ in one round is merely temporarily hidden from the LLM rather than permanently deleted.
If a later sub-question redirects the search toward its neighborhood, that triple can reappear in a future potential subgraph, allowing the agent to recover from earlier pruning mistakes.
At the same time, to maximize exploration efficiency, triples that have been successfully absorbed into $\mathcal{G}^{\mathrm{mem}}$ are excluded from redundant future expansions, whereas triples explicitly rejected by the LLM are permanently masked from subsequent retrieval.

\subsection{Complexity Analysis}
Let $L_t$ denote the number of transformer blocks used in the question-guided triple encoder and $L_m$ denote the number of
foresight feedback propagation layers.
Let $|V|$ and $|E|$ denote the number of entities and triple instances in the expanded subgraph
$\mathcal{G}^{\mathrm{xpd}}$, and let $d$ be the representation dimension and we assume $d_h=d$ for simplicity.
During layered BFS expansion,
FoG scores incident triples on frontier entities; 
let $M$ denote the total number of
candidate triple instances scored in this process.

Since each triple is encoded with a constant number of tokens, the per-triple computation in a transformer block is dominated by $O(d^2)$. Therefore, encoding at most $M$ triples with $L_t$ blocks incurs a worst-case cost of $O(L_t M d^2)$.
For the foresight triple scorer, computing attention over $S$ sub-questions and the final MLP score costs $O(Sd^2)$ per
scored triple, yielding $O(SMd^2)$ during expansion, and $O(S\,|E|\,d^2)$ when re-scoring all edges after propagation.
For each message passing layer, edge-wise message computation and attention aggregation are linear in edges and nodes, giving
$O((|E|+|V|)\,d^2)$ per layer, hence $O(L_m(|E|+|V|)\,d^2)$ for $L_m$ layers.
Combining the above, the overall graph-side time complexity is
$
O\Big(\big(L_t M + S(M+|E|) + L_m(|E|+|V|)\big)\,d^2\Big).
$


\begin{table*}[t]
\caption{Overall results on WebQSP and CWQ. The $\dagger$ and $\ddagger$ symbols denote graph retrieval-based methods and LLM-guided graph agents method, respectively. The \colorbox{blue!10}{$\star$} indicates the best-performing baseline. The best and second-best results are highlighted in \textbf{bold} and \underline{underlined}, respectively.}
\centering
\footnotesize
\renewcommand{\arraystretch}{0.6}

\setlength{\tabcolsep}{2pt} 
\begin{tabular*}{\textwidth}{
@{\extracolsep{\fill}}
l l l c c c c c c
@{}
}
\toprule
\multirow{2}[2]{*}{\textbf{Type}} & \multirow{2}[2]{*}{\textbf{Method}} & \multirow{2}[2]{*}{\textbf{LLM}} & \multicolumn{3}{c}{\textbf{WebQSP}} & \multicolumn{3}{c}{\textbf{CWQ}} \\
\cmidrule(lr){4-6} \cmidrule(lr){7-9}
 & & & \textbf{Hit} & \textbf{Hit@1} & \textbf{F1} & \textbf{Hit} & \textbf{Hit@1} & \textbf{F1} \\
\midrule

\multirow{4}{*}{\begin{tabular}{@{}l@{}}Non-LLM\end{tabular}} 
 & NSM            & - & - & 74.30 & - & - & 48.80 & - \\
 & SQALER         & - & - & 76.10 & - & - & -    & - \\
 & KGT5           & - & - & 56.10 & - & - & 36.50 & - \\
 & UniKGQA        & - & - & 77.20 & 72.20 & - & 51.20 & 49.40 \\
\midrule

\multirow{3}{*}{\begin{tabular}{@{}l@{}}LLM-only\end{tabular}} 
 & Zero-shot      & GPT-4o & 58.24 & 58.24 & 42.57 & 35.82 & 35.82 & 31.49 \\
 & Few-shot       & GPT-4o & 64.13 & 64.13 & 46.75 & 43.30 & 43.30 & 39.58 \\
 & CoT            & GPT-4o & 66.09 & 66.09 & 52.95 & 46.97 & 46.97 & 42.71 \\
\midrule

\multirow{5}{*}{\raisebox{-2.8ex}{\begin{tabular}{@{}l@{}}Fine-tuned\\LLM\end{tabular}}}
 & G-Retriever$^{\dagger}$    & LLaMA2-7B & - & 70.10 & -    & - & -    & - \\
 & GNN-RAG$^{\dagger}$        & LLaMA2-7B & {\setlength{\fboxsep}{1pt}\colorbox{blue!10}{90.70$^\star$}} & 82.80 & 73.50 & 68.70 & 62.80 & 60.40 \\
  & LightPROF$^{\dagger}$      & LLaMA3-8B & - & 83.77 & -    & - & 59.26 & - \\
 & EPERM$^{\ddagger}$         & LLaMA2-7B & -    & 88.80 & 72.40 & -    & 66.20 & 58.90 \\
 & KG-Agent$^{\ddagger}$      & LLaMA2-7B & -    & 83.30 & {\setlength{\fboxsep}{1pt}\colorbox{blue!10}{\underline{81.00$^\star$}}} & - & 72.20 & {\setlength{\fboxsep}{1pt}\colorbox{blue!10}{\underline{69.80$^\star$}}} \\

\midrule

\multirow{11}{*}{\begin{tabular}{@{}l@{}}In-context\\LLM\end{tabular}} 
  & Subgraph-RAG$^{\dagger}$   & GPT-4o & 88.26 & 84.73 & 74.98 & 60.36 & 56.39 & 52.40 \\
 & BYOKG-RAG$^{\ddagger}$     & Claude-3.5 & 86.60 & - & - & 73.60 & - & - \\
 & ORT$^{\ddagger}$           & DeepSeek-v3 & - & {\setlength{\fboxsep}{1pt}\colorbox{blue!10}{89.43$^\star$}} & 71.83 & - & {\setlength{\fboxsep}{1pt}\colorbox{blue!10}{72.91$^\star$}} & 62.63 \\
 & ToG$^{\ddagger}$           & GPT-4 & 82.60 & - & 36.40 & 69.50 & - & 31.80 \\
 & RoG$^{\ddagger}$           & GPT-4 & 85.70 & 80.80 & 70.80 & 62.60 & 57.80 & 56.20 \\
 & PoG$^{\ddagger}$           & GPT-4 & 87.30 & - & - & {\setlength{\fboxsep}{1pt}\colorbox{blue!10}{75.00$^\star$}} & - & - \\
 & FiDeLis$^{\ddagger}$       & GPT-4 & - & 84.39 & 78.32 & - & 71.47 & 64.32 \\
 
 \noalign{\vskip -0.3ex}
\cmidrule(l{1pt}r{1pt}){2-9}
\noalign{\vskip -0.3ex}

 & \textbf{FoG}   & GPT-4o-mini & 88.68$_{\pm 0.13}$ & 87.23$_{\pm 0.18}$ & 75.16$_{\pm 0.14}$ & 82.33$_{\pm 0.11}$ & 73.45$_{\pm 0.09}$ & 66.56$_{\pm 0.15}$ \\
 & \textbf{FoG}   & GPT-4       & \underline{91.31}$_{\pm 0.10}$ & \textbf{90.17}$_{\pm 0.18}$ & 79.57$_{\pm 0.17}$ & \textbf{87.44}$_{\pm 0.14}$ & \underline{78.39}$_{\pm 0.17}$ & 68.50$_{\pm 0.16}$ \\
 & \textbf{FoG}   & GPT-4o      & \textbf{91.63}$_{\pm 0.15}$ & \underline{89.66}$_{\pm 0.18}$ & \textbf{81.28}$_{\pm 0.19}$ & \underline{87.03}$_{\pm 0.13}$ & \textbf{80.21}$_{\pm 0.19}$ & \textbf{71.34}$_{\pm 0.16}$ \\
\bottomrule
\end{tabular*}
\label{tab:main-result}
\end{table*}

\section{Experiments}

\subsection{Experimental Settings}
\label{sec:experimental-settings}

\noindent\textbf{Datasets.}
We evaluate FoG on two widely used Freebase-based KBQA benchmarks, WebQSP~\cite{webqsp} and CWQ~\cite{cwq}, using their official train/test splits. Both datasets require multi-hop reasoning over structured knowledge bases. As shown in Figure~\ref{fig:degree}, CWQ has more high-degree nodes than WebQSP, resulting in a larger branching factor and more challenging graph exploration. Further details are provided in Appendix~\ref{sec:dataset_descriptions}.

\noindent\textbf{Metrics.}
We use Hit, Hit@1, and F1 to evaluate whether the prediction contains a correct answer, whether the top-ranked answer is correct, and the precision--recall trade-off, respectively.

\noindent\textbf{Baselines.}
We compare FoG against 19 baselines, grouped into embedding-based, prompting-based, fine-tuned LLM, and in-context LLM methods.
Descriptions of the baselines are provided in Appendix~\ref{sec:app-baselines}.

\noindent\textbf{Implementation Details.}
We adopt Qwen3-Embedding-4B \cite{zhang2025qwen3} as a unified pretrained encoder to embed both natural-language queries and KG symbols (entities/relations). The encoder is kept frozen throughout training and inference.
We deploy the original Freebase as the underlying knowledge base and evaluate our method on WebQSP and CWQ \cite{webqsp,cwq}.
For foresight subgraph expansion, we set the hop budget to $L{=}3$ for all experiments. We use an expansion threshold $\tau_e{=}0.1$ to select frontier triples during FSE, and a potential threshold $\tau_p{=}0.7$ to form the potential subgraph passed to the memory subgraph manager.

\subsection{Main Results}

Table~\ref{tab:main-result} presents the overall results on WebQSP and CWQ. 
Without task-specific fine-tuning, FoG consistently outperforms prior fine-tuned LLM methods as well as strong graph retrieval-based and LLM-guided graph agent baselines. 
It achieves relative Hit@1 improvements of 1.02\% on WebQSP and 16.58\% on CWQ over the strongest baselines, with larger gains on CWQ highlighting its effectiveness for compositional questions with larger branching factors.
This is mainly because FoG propagates distant relevance signals back to near-source candidates while maintaining a compact memory subgraph.

\subsection{Ablation Study}

\begin{table*}[t]
\caption{
Ablation results on WebQSP and CWQ.
FSE, FFP, and MSM denote Foresight Subgraph Expansion, Foresight Feedback Propagation, and Memory Subgraph Manager, respectively.
}
\centering
\vspace{-0.15cm}
\footnotesize
\renewcommand{\arraystretch}{1}
\setlength{\tabcolsep}{1.5pt}
\begin{tabular*}{\textwidth}{@{\extracolsep{\fill}}lcccccccc@{\hspace{0.9cm}}}
\toprule
& \multicolumn{4}{c}{\textbf{WebQSP}} & \multicolumn{4}{c}{\textbf{CWQ}} \\
\cmidrule(lr){2-5}\cmidrule(lr){6-9}
\textbf{Method} 
& \textbf{Hit} & \textbf{Hit@1} & \textbf{Hit@5} & \textbf{F1} 
& \textbf{Hit} & \textbf{Hit@1} & \textbf{Hit@5} & \textbf{F1} \\
\midrule

FoG w/o FSE
& 90.42\rlap{$_{\textcolor{green!50!black}{\downarrow 1.21}}$}
& 88.78\rlap{$_{\textcolor{green!50!black}{\downarrow 0.88}}$}
& 90.19\rlap{$_{\textcolor{green!50!black}{\downarrow 1.38}}$}
& 77.41\rlap{$_{\textcolor{green!50!black}{\downarrow 3.87}}$}
& 85.22\rlap{$_{\textcolor{green!50!black}{\downarrow 1.81}}$}
& 73.95\rlap{$_{\textcolor{green!50!black}{\downarrow 6.26}}$}
& 82.70\rlap{$_{\textcolor{green!50!black}{\downarrow 2.10}}$}
& 62.98\rlap{$_{\textcolor{green!50!black}{\downarrow 8.36}}$} \\

FoG w/o FFP
& 89.27\rlap{$_{\textcolor{green!50!black}{\downarrow 2.36}}$}
& 86.92\rlap{$_{\textcolor{green!50!black}{\downarrow 2.74}}$}
& 88.74\rlap{$_{\textcolor{green!50!black}{\downarrow 2.83}}$}
& 73.85\rlap{$_{\textcolor{green!50!black}{\downarrow 7.43}}$}
& 78.61\rlap{$_{\textcolor{green!50!black}{\downarrow 8.42}}$}
& 68.43\rlap{$_{\textcolor{green!50!black}{\downarrow 11.78}}$}
& 76.02\rlap{$_{\textcolor{green!50!black}{\downarrow 8.78}}$}
& 57.34\rlap{$_{\textcolor{green!50!black}{\downarrow 14.00}}$} \\

FoG w/o MSM
& 90.19\rlap{$_{\textcolor{green!50!black}{\downarrow 1.44}}$}
& 88.89\rlap{$_{\textcolor{green!50!black}{\downarrow 0.77}}$}
& 90.04\rlap{$_{\textcolor{green!50!black}{\downarrow 1.53}}$}
& 77.19\rlap{$_{\textcolor{green!50!black}{\downarrow 4.09}}$}
& 79.73\rlap{$_{\textcolor{green!50!black}{\downarrow 7.30}}$}
& 67.12\rlap{$_{\textcolor{green!50!black}{\downarrow 13.09}}$}
& 75.84\rlap{$_{\textcolor{green!50!black}{\downarrow 8.96}}$}
& 56.15\rlap{$_{\textcolor{green!50!black}{\downarrow 15.19}}$} \\

\textbf{FoG}
& \textbf{91.63}
& \textbf{89.66}
& \textbf{91.57}
& \textbf{81.28}
& \textbf{87.03}
& \textbf{80.21}
& \textbf{84.80}
& \textbf{71.34} \\

\bottomrule
\end{tabular*}
\vspace{-0.2cm}
\label{tab:ablation-study}
\end{table*}

\begin{figure*}[t]
    \centering
    \vspace{3pt}

    \begin{minipage}[t]{0.375\textwidth}
        \vspace{0pt}
        \centering
        \includegraphics[
            width=\linewidth,
            height=0.17\textheight,
            keepaspectratio
        ]{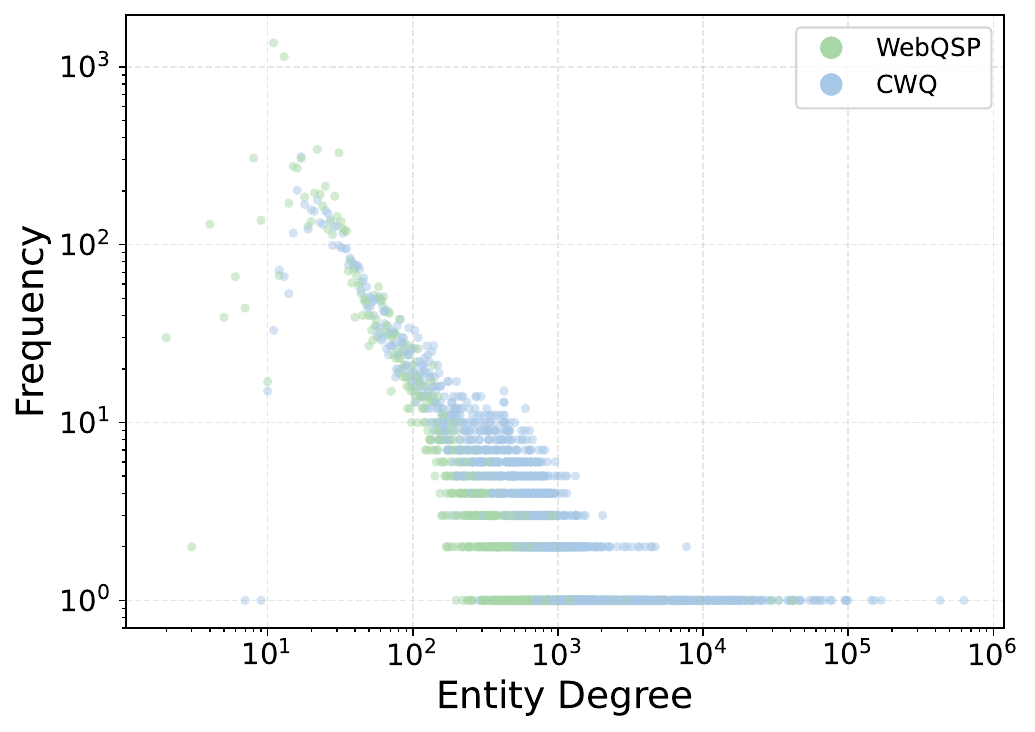}
        \captionof{figure}{Degree distribution of Freebase entities referenced in CWQ and WebQSP.}
        \label{fig:degree}
    \end{minipage}
    \hfill
    \begin{minipage}[t]{0.585\textwidth}
        \vspace{0pt}
        \centering

        \begin{subfigure}[t]{0.49\linewidth}
            \centering
            \includegraphics[
                width=\linewidth,
                height=0.17\textheight,
                keepaspectratio
            ]{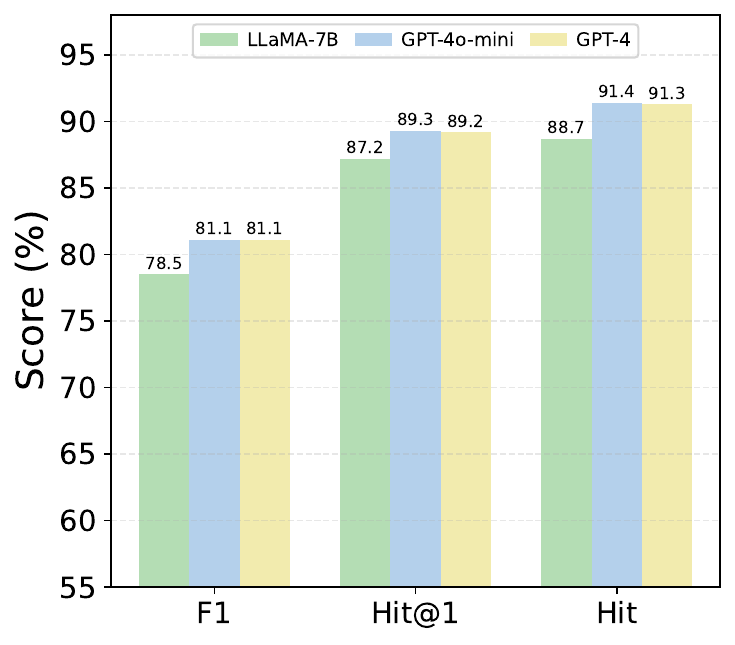}
            \caption{WebQSP.}
            \label{fig:sub-ques-webqsp}
        \end{subfigure}
        \hspace{-0.003\linewidth}
        \begin{subfigure}[t]{0.49\linewidth}
            \centering
            \includegraphics[
                width=\linewidth,
                height=0.17\textheight,
                keepaspectratio
            ]{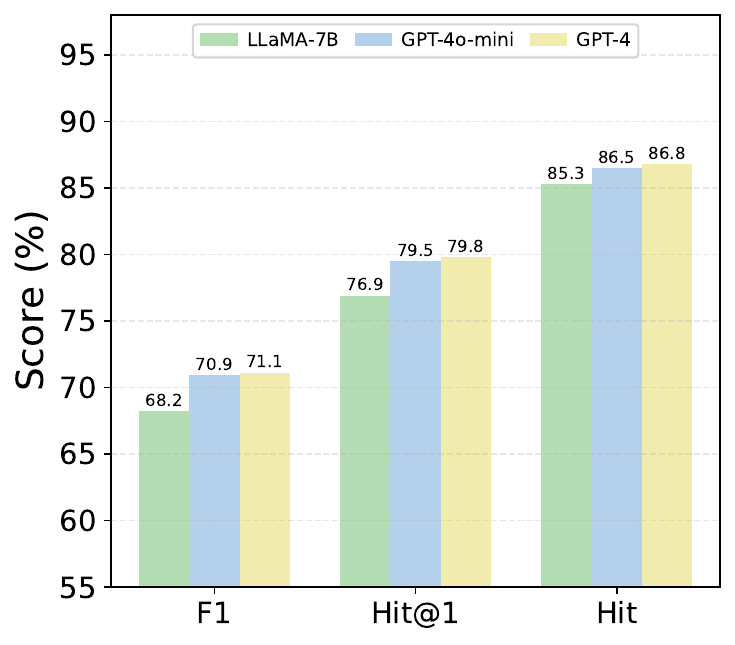}
            \caption{CWQ.}
            \label{fig:sub-ques-cwq}
        \end{subfigure}

        \captionof{figure}{Performance comparison of different sub-question LLMs.}
        \label{fig:sub-ques}
    \end{minipage}

\end{figure*}


\noindent\textbf{Effect of Core Components.}
Table~\ref{tab:ablation-study} reports ablation results on WebQSP and CWQ to illustrate the contribution of each component. Removing Foresight Feedback Propagation leads to the most pronounced degradation, showing that far-to-near feedback is central to refining multi-hop evidence and improving pruning decisions. The Memory Subgraph Manager is particularly important on the more compositional CWQ benchmark, suggesting that maintaining accumulated evidence is crucial for deeper iterative reasoning. Removing Foresight Subgraph Expansion also causes consistent performance drops, indicating that high-recall initial expansion provides useful candidate evidence for subsequent feedback and memory updates. Overall, the results show that the three modules are complementary and work best in combination.

\noindent\textbf{LLM Flexibility Study.}
Table~\ref{tab:main-result} reports results with different backbone LLMs to demonstrate that FoG is not reliant on any particular LLM. Specifically, we consider three LLMs: LLaMA-7B, GPT-4o-mini and GPT-4. The results show that FoG consistently outperforms other baselines across different backbones and achieves substantial gains in most metrics, demonstrating robust effectiveness and strong transferability.

\begin{wraptable}{r}{0.50\textwidth}
\caption{Comparison between standard GNN propagation and FFP.}
\centering
\small
\renewcommand{\arraystretch}{1}
\setlength{\tabcolsep}{1pt}
\begin{tabular*}{\linewidth}{@{\extracolsep{\fill}}lcccccc@{}}
\toprule
\multirow{2}{*}{\textbf{Method}} 
& \multicolumn{3}{c}{\textbf{WebQSP}} 
& \multicolumn{3}{c}{\textbf{CWQ}} \\
\cmidrule(lr){2-4} \cmidrule(lr){5-7}
& \textbf{Hit} & \textbf{Hit@1} & \textbf{F1} 
& \textbf{Hit} & \textbf{Hit@1} & \textbf{F1} \\
\midrule
FoG w/ GNN & 89.92 & 88.13 & 78.75 & 84.06 & 76.79 & 67.27 \\
FoG w/ FFP & 91.63 & 89.66 & 81.28 
           & 87.03 & 80.21 & 71.34 \\
\bottomrule
\end{tabular*}
\label{tab:gnn_ablation}
\end{wraptable}
\noindent\textbf{Comparison with Standard GNN.}
To test whether FoG's gains come simply from generic message passing, we replace Foresight Feedback Propagation (FFP) with a standard GNN on the same expanded subgraph, treating the graph as undirected and using the same edge-level supervision. As shown in Table~\ref{tab:gnn_ablation}, this variant consistently underperforms FFP on both WebQSP and CWQ, indicating that generic neighborhood aggregation is insufficient for foresight-aware pruning. Unlike undirected GNN propagation, which can blur the distinction between relevant triples and distractors, FFP propagates signals from distant frontier nodes back to source-side candidates, better matching FoG's early-pruning bottleneck.

\noindent\textbf{Sensitivity to Sub-question Decomposition.}
FoG uses an LLM to decompose the original question into sub-questions, which provide high-level guidance for graph exploration. Since this may introduce dependency on the sub-question generator, we replace it with LLMs of different strengths while keeping the graph retriever, foresight feedback propagation module, memory subgraph manager, and final answer generation unchanged.
As shown in Figure~\ref{fig:sub-ques}, stronger sub-question LLMs generally improve performance on both WebQSP and CWQ, as better decompositions can guide FoG toward more informative graph regions. Nevertheless, the overall trend remains stable, and lightweight models still achieve competitive results across benchmarks. This indicates that FoG is not merely relying on a powerful LLM; rather, its foresight-aware retrieval and memory subgraph mechanisms provide a robust reasoning backbone.

\begin{wraptable}{r}{0.50\textwidth}
\caption{Efficiency comparison on WebQSP and CWQ: average LLM calls and token usage per question.}
\centering
\vspace{-0.15cm}   
\small
\renewcommand{\arraystretch}{1}
\setlength{\tabcolsep}{2pt}
\begin{tabular*}{\linewidth}{@{\extracolsep{\fill}}lccccc@{}}
\toprule
\textbf{Dataset} & \textbf{Method} & \textbf{Call} & \textbf{Input} & \textbf{Output} & \textbf{Total} \\
\midrule
\multirow{3}{*}{WebQSP} & ToG & 15.9 & 6031.2 & 987.7 & 7018.9 \\
                        & PoG & 9.0  & 5234.8 & 282.9 & 5517.7 \\
                        & FoG & \textbf{2.3}  & \textbf{1837.3} & \textbf{248.9} & \textbf{2086.2} \\
\hline
\multirow{3}{*}{CWQ}    & ToG & 22.6 & 8182.9 & 1486.4 & 9669.4 \\
                        & PoG & 13.3 & 7803.0 & 353.2  & 8156.2 \\
                        & FoG & \textbf{2.8}  & \textbf{3461.7} & \textbf{295.4}  & \textbf{4585.1} \\
\bottomrule
\end{tabular*}

\label{tab:eff-analysis}
\end{wraptable}

\subsection{Efficiency Analysis}
FoG requires substantially fewer LLM calls than LLM-guided graph agents baselines.
Unlike methods such as ToG that perform beam-style graph search, FoG maintains a memory subgraph and avoids expanding and scoring $K$ partial paths at each depth.
For ToG with search depth $L$ and beam width $K$, the number of LLM calls scales on the order of $L\!\cdot\!K$.
In contrast, FoG shifts multi-hop expansion and edge filtering to the graph side and uses the LLM only for high-level updates on a compressed potential subgraph together with the memory subgraph.
With $R$ reasoning rounds, FoG uses roughly $R{+}1$ LLM calls per question, independent of $K$.
This design also reduces token usage since the LLM does not need to read all candidate triples.
Although FoG introduces additional graph-side computation for subgraph expansion, triple scoring, and feedback propagation, these operations are lightweight and parallelizable; in our implementation, their overhead is minor compared with the latency and cost of LLM inference.
Table~\ref{tab:eff-analysis} reports detailed statistics.



\begin{figure}[t]
  \centering
  \begin{subfigure}[t]{0.24\textwidth}
    \centering
    \includegraphics[width=\linewidth]{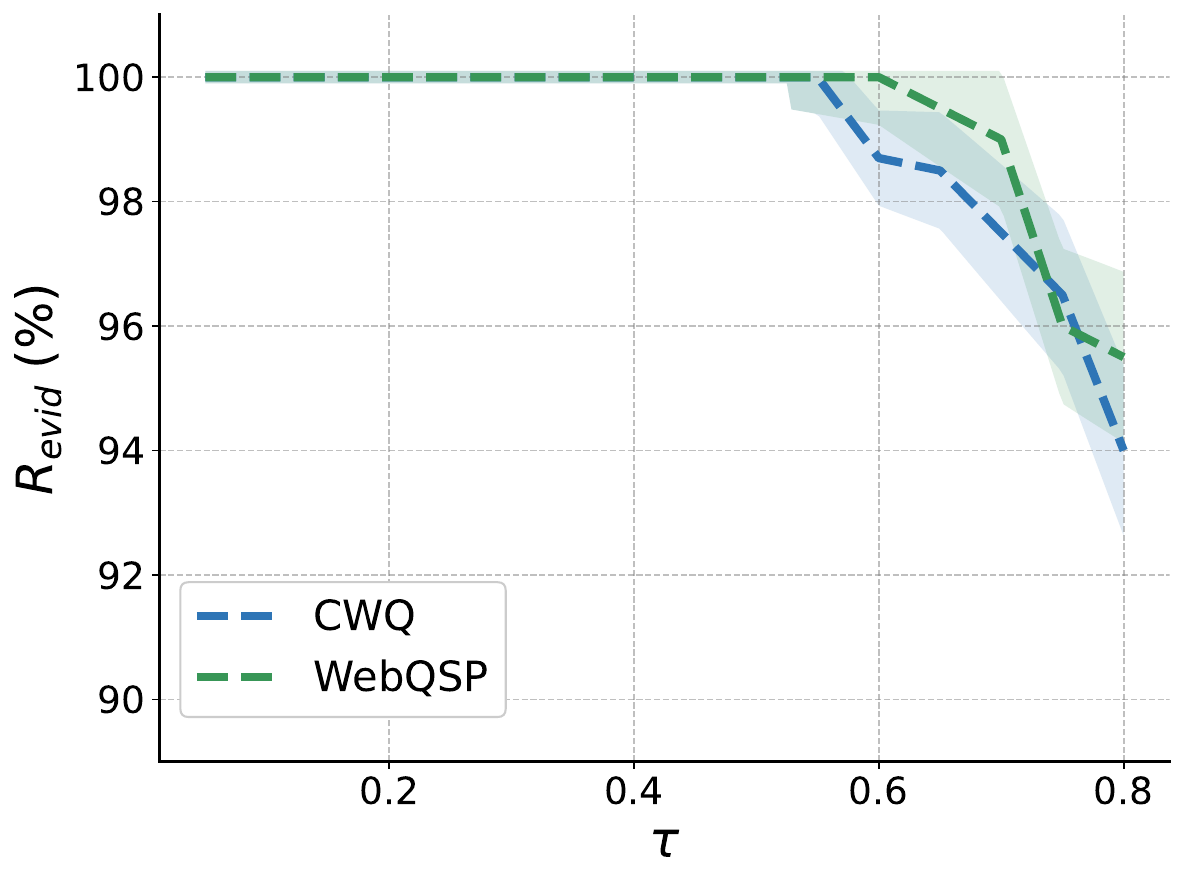}
    \caption{Evidence recall $R_{\text{evid}}$}
    \label{fig:retr-cap-a}
  \end{subfigure}
  \hfill
  \begin{subfigure}[t]{0.24\textwidth}
    \centering
    \includegraphics[width=\linewidth]{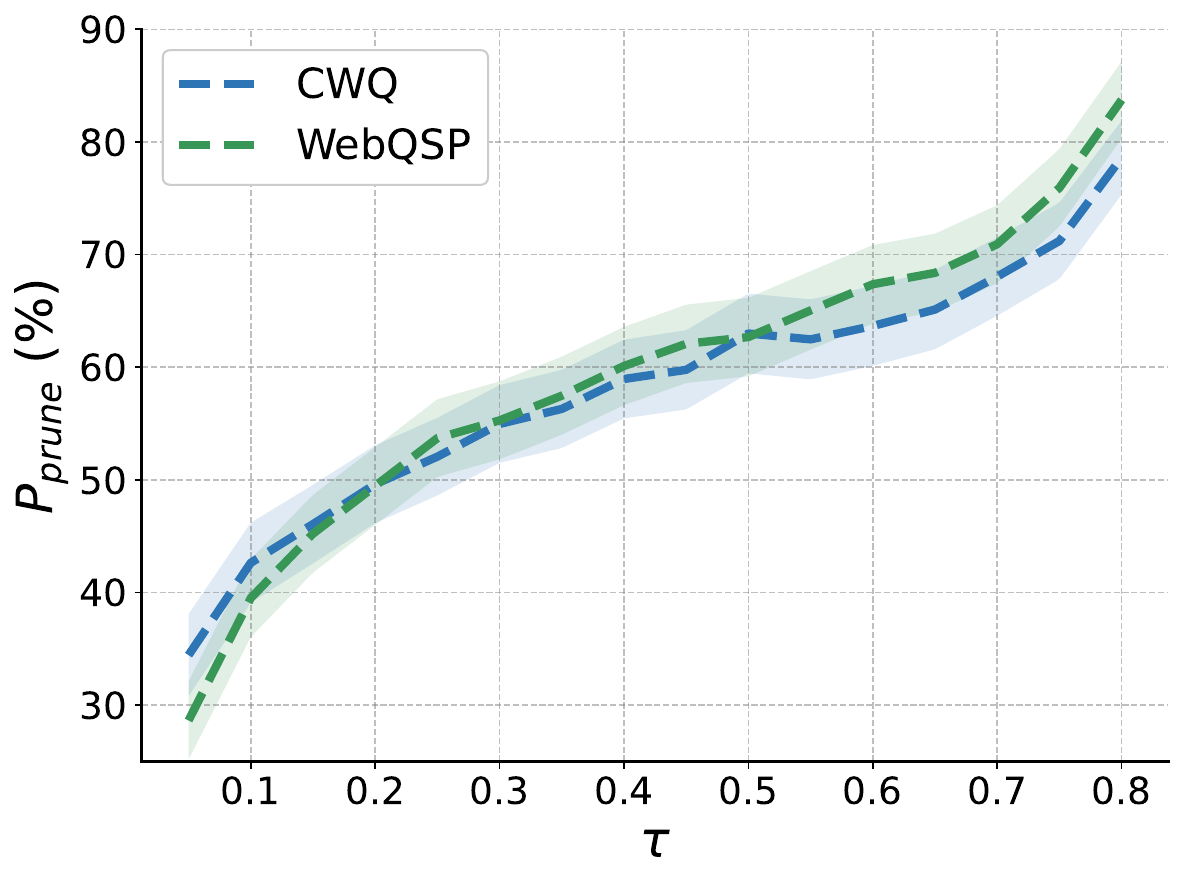}
    \caption{Noise pruning $P_{\text{prune}}$}
    \label{fig:retr-cap-b}
  \end{subfigure}
  \hfill
  \begin{subfigure}[t]{0.24\textwidth}
    \centering
    \includegraphics[width=\linewidth]{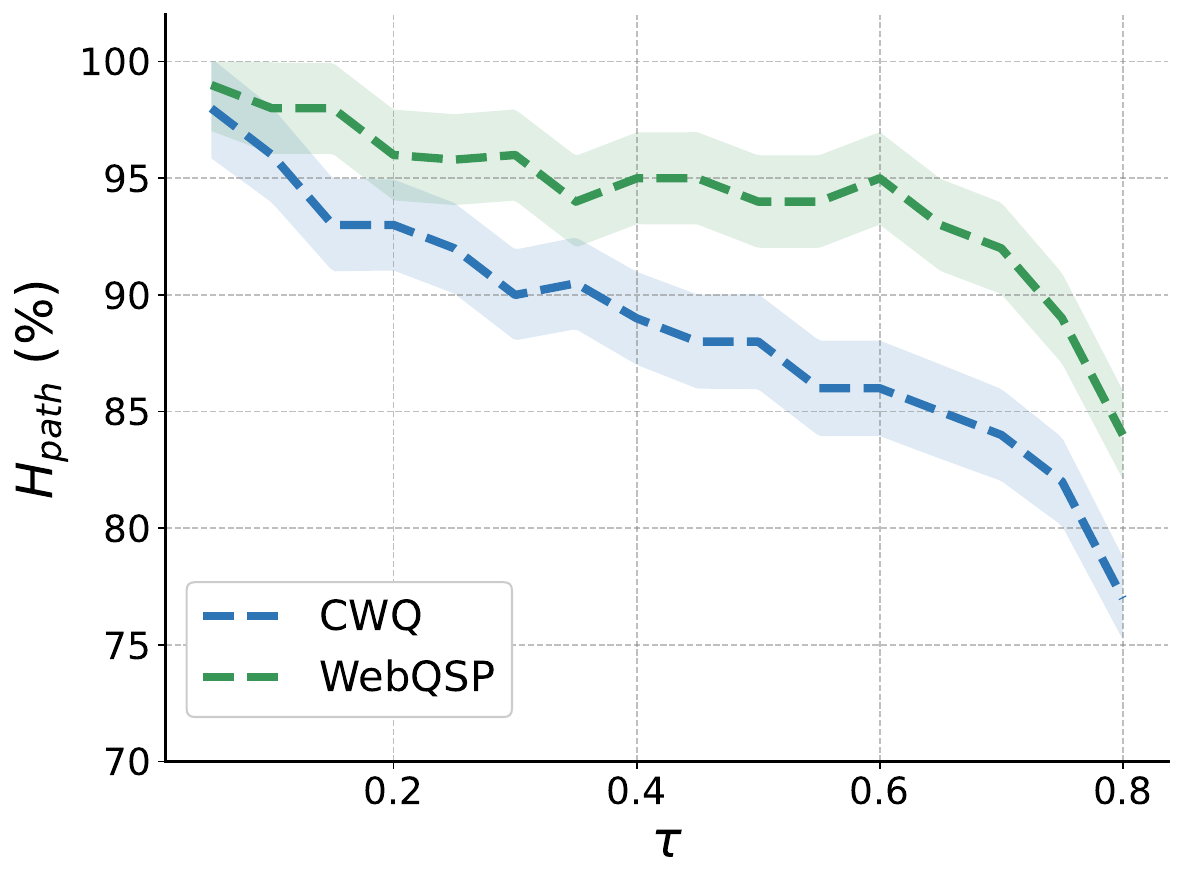}
    \caption{Path hit $H_{\text{path}}$}
    \label{fig:retr-cap-c}
  \end{subfigure}
  \hfill
  \begin{subfigure}[t]{0.24\textwidth}
    \centering
    \includegraphics[width=\linewidth]{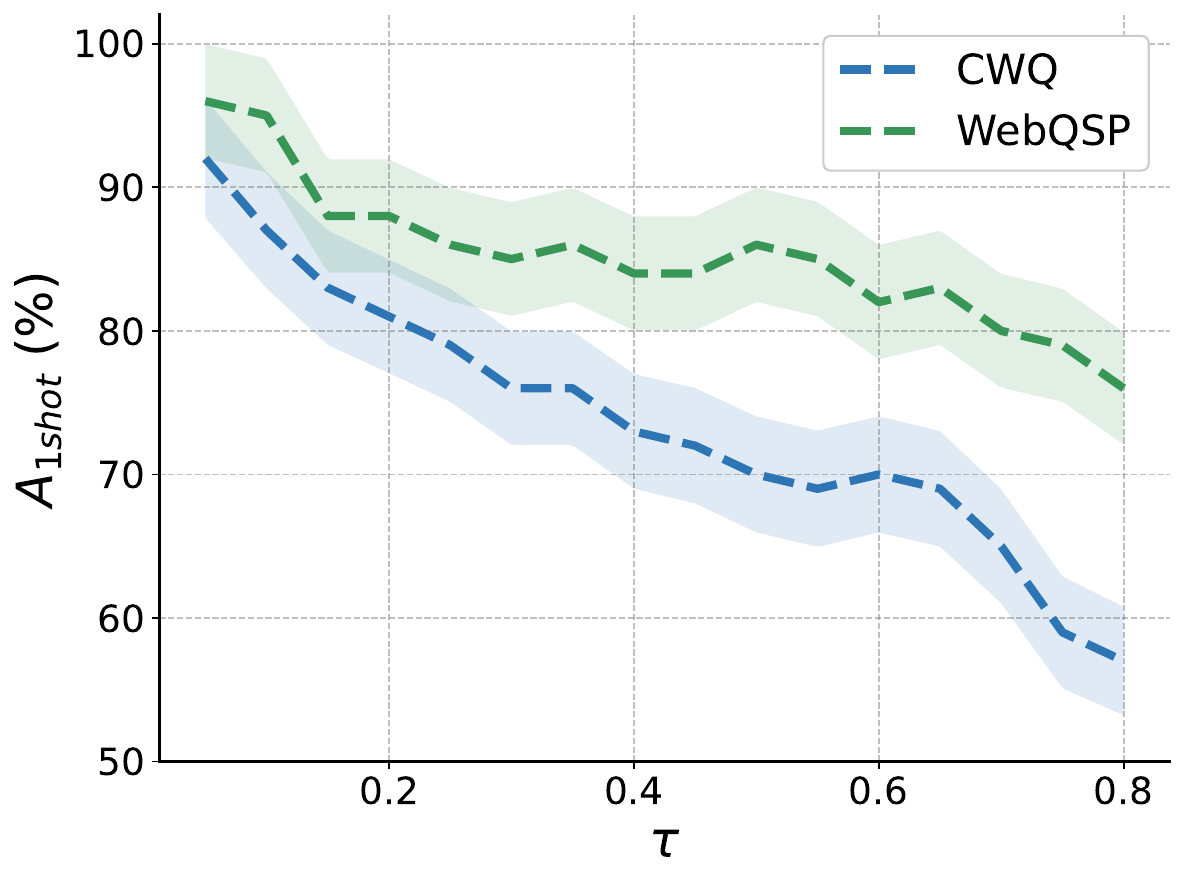}
    \caption{One-shot rate $A_{\text{1shot}}$}
    \label{fig:retr-cap-d}
  \end{subfigure}
  \caption{Effect of $\tau_p$ on knowledge retrieval capacity (CWQ and WebQSP).}
  \label{fig:retr-cap}
\end{figure}

\subsection{Knowledge Retrieval Capacity}

We study how the retrieval thresholds affect FoG's knowledge retrieval capacity by sweeping them over $[0.05, 0.90]$ on WebQSP and CWQ. Specifically, FoG uses two threshold-based filtering steps: the expansion threshold $\tau_e$ filters scored frontier triples during subgraph expansion, while the potential threshold $\tau_p$ filters re-scored triples before constructing the potential subgraph. Although $\tau_e$ and $\tau_p$ are applied at different stages, both act as score thresholds for retaining question-relevant triples.

As shown in Figure~\ref{fig:retr-cap}, the thresholds govern a clear trade-off between evidence recall and noise pruning. Larger thresholds make FoG more selective, removing more irrelevant triples and improving $P_{\text{prune}}$, but they may also filter out useful evidence and reduce $R_{\text{evid}}$. With moderate threshold values, $H_{\text{path}}$ remains high, indicating that FoG usually preserves at least one correct evidence path. However, retrieving the complete evidence set in a single round is harder, and $A_{\text{1shot}}$ decreases more quickly, especially on CWQ.

\subsection{Case study}
We conduct a qualitative analysis to better understand the typical failure modes of existing graph-guided reasoning methods and how FoG mitigates them.

\begin{wrapfigure}{r}{0.38\textwidth}
\vspace{-10pt}
\centering
\includegraphics[width=\linewidth]{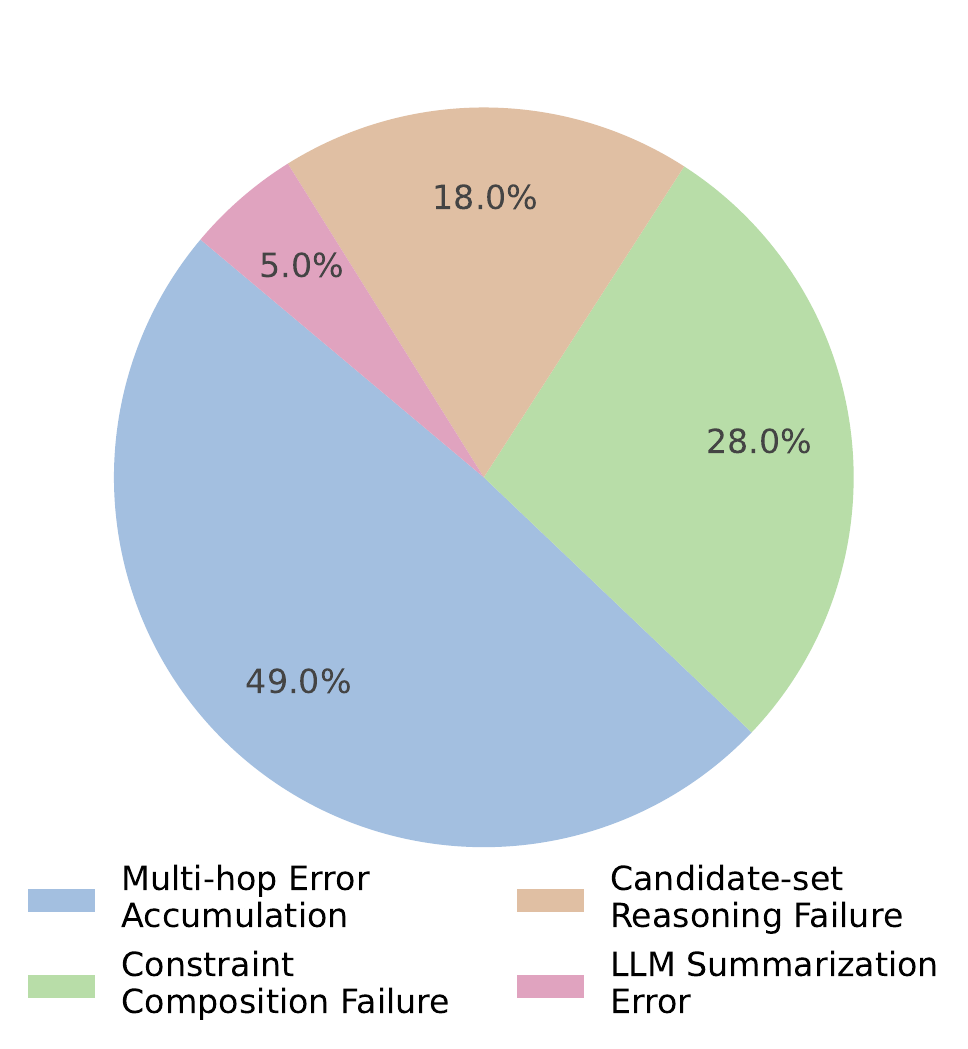}
\vspace{-0.2cm}
\caption{Error distribution of existing graph-guided reasoning failures.}
\label{fig:error_pie}
\vspace{-0.2cm}
\end{wrapfigure}
\noindent\textbf{Error Analysis of Existing Methods.}
We manually inspect failed predictions of graph-guided reasoning methods and group them into four error types, as shown in Figure~\ref{fig:error_pie}. The dominant failure is \emph{Multi-hop Error Accumulation}, where early relation or entity selection errors shift the search frontier and make the correct evidence chain hard to recover. Other failures include \emph{Constraint Composition Failure}, where only part of the required constraints are verified; \emph{Candidate-set Reasoning Failure}, where comparison or ranking over retrieved candidates is needed but missing; and \emph{LLM Summarization Error}, where the correct answer is retrieved but missed due to label variation or imperfect normalization.

\noindent\textbf{Qualitative Case Study.}
We further present a representative case to illustrate how short-sighted graph exploration loses answer-critical evidence. When the correct evidence lies several hops away, local scoring may prune a weak-looking but crucial relation. FoG mitigates this by using foresight signals to preserve globally useful branches and recover the correct reasoning path. Detailed examples are provided in Appendix~\ref{app:case-study}.
\section{Related Work}
\label{sec:related_work}

Complex KBQA requires composing multi-hop evidence from large structured knowledge bases. Recent studies explore grounding LLM reasoning in such structured knowledge~\cite{huang2024promptinglongtail,linders2025kgrag,lee2025hybgrag, xu2025noderag, chen2026pathrag}.
 
One line of work focuses on \emph{graph-embedding based retrieval}, which retrieves, ranks, or encodes compact graph evidence before downstream reasoning \cite{jiang2023structgpt, zhang2024knowgpt, he2024gretriever, li2025simple, mavromatis2025gnnrag, ao2025lightprof}. These methods improve grounding by reducing large knowledge bases into smaller evidence contexts. However, they mainly judge whether an edge or triple currently appears relevant to the question, rather than whether the branch opened by this edge can eventually lead to the true evidence chain. Thus, locally ambiguous but globally necessary triples may still be pruned before their downstream value becomes visible.

Another line of work studies \emph{LLM-guided graph agents}, where the LLM plans, traverses, verifies, or revises reasoning paths over knowledge graphs \cite{wen2024mindmap, thinkongraph, luo2023rog, chen2024pog, long2025eperm, sui2025fidelis, mavromatis2025byokgrag}. These methods enable adaptive exploration, but they usually formulate reasoning as sequential graph search. Once an answer-critical relation is excluded from the search space or the search shifts to an incorrect branch, later steps have limited ability to recover the correct path.

FoG takes a different perspective on graph retrieval and reasoning. Compared with graph-based retrieval methods, which ask whether an edge currently looks relevant to the question, FoG asks whether the branch opened by this edge can eventually lead to answer-supporting evidence. It therefore evaluates early triples by their downstream utility rather than only by their immediate relevance. Compared with LLM-guided graph agents, which may suffer from short-sighted pruning during sequential traversal, FoG anticipates the value of early branches before committing to a reasoning path. By propagating distant relevance signals back to near-source candidates and maintaining a compact memory subgraph, FoG reduces myopic pruning errors while preserving globally useful evidence for complex KBQA.

\section{Conclusion \& Limitation}
\label{sec:conclusion-and-limitation}
This paper proposed Foresight-over-Graph (FoG) to reduce short-sighted pruning in multi-hop KBQA by injecting future-aware signals into current graph retrieval. With foresight feedback propagation and a memory subgraph manager, FoG preserves crucial evidence while exposing only compressed subgraphs to the LLM. Experiments on WebQSP and CWQ show that FoG outperforms prior methods in accuracy and delivers markedly better efficiency across different LLM backbones. \textbf{Limitations:} Although FoG has been evaluated on two widely used KGQA benchmarks, further validation on more diverse and domain-specific knowledge graphs, such as medical or legal QA, would strengthen its generalizability, which we leave for future work.

\bibliography{custom}

\appendix

\section*{NeurIPS Paper Checklist}

\begin{enumerate}

\item {\bf Claims}
    \item[] Question: Do the main claims made in the abstract and introduction accurately reflect the paper's contributions and scope?
    \item[] Answer: \answerYes{} 
    \item[] Justification: Our main claim that complex QA requires foresight-aware retrieval is both methodologically and empirically supported by FoG’s design and results on WebQSP/CWQ.

\item {\bf Limitations}
    \item[] Question: Does the paper discuss the limitations of the work performed by the authors?
    \item[] Answer: \answerYes{}{} 
    \item[] Justification: Please refer to Section~
\ref{sec:conclusion-and-limitation}.

\item {\bf Theory assumptions and proofs}
    \item[] Question: For each theoretical result, does the paper provide the full set of assumptions and a complete (and correct) proof?
    \item[] Answer: \answerNA{} 
    \item[] Justification: This paper does not include theoretical results.

    \item {\bf Experimental result reproducibility}
    \item[] Question: Does the paper fully disclose all the information needed to reproduce the main experimental results of the paper to the extent that it affects the main claims and/or conclusions of the paper (regardless of whether the code and data are provided or not)?
    \item[] Answer: \answerYes{} 
    \item[] Justification: We have thoroughly provided the relevant information in section~\ref{sec:experimental-settings}

\item {\bf Open access to data and code}
    \item[] Question: Does the paper provide open access to the data and code, with sufficient instructions to faithfully reproduce the main experimental results, as described in supplemental material?
    \item[] Answer: \answerYes{}{} 
    \item[] Justification: We have provided the relevant codes with clear instructions in an anonymous link (see Abstract). All the datasets we used are puplicly available and clearly cited.

\item {\bf Experimental setting/details}
    \item[] Question: Does the paper specify all the training and test details (e.g., data splits, hyperparameters, how they were chosen, type of optimizer) necessary to understand the results?
    \item[] Answer: \answerYes{}{} 
    \item[] Justification: We have described the dataset setups and evaluation metrics in Section~\ref{sec:experimental-settings}

\item {\bf Experiment statistical significance}
    \item[] Question: Does the paper report error bars suitably and correctly defined or other appropriate information about the statistical significance of the experiments?
    \item[] Answer: \answerYes{}{} 
    \item[] Justification: The experimental results are the average of three runs to avoid the random bias.

\item {\bf Experiments compute resources}
    \item[] Question: For each experiment, does the paper provide sufficient information on the computer resources (type of compute workers, memory, time of execution) needed to reproduce the experiments?
    \item[] Answer: \answerYes{}{} 
    \item[] Justification: The paper reports the GPU used and provides efficiency statistics on LLM calls and token usage. Please refer to Section~\ref{sec:experimental-settings}
    
\item {\bf Code of ethics}
    \item[] Question: Does the research conducted in the paper conform, in every respect, with the NeurIPS Code of Ethics \url{https://neurips.cc/public/EthicsGuidelines}?
    \item[] Answer: \answerYes{} 
    \item[] Justification: The research was conducted in accordance with the NeurIPS Code of Ethics, and we identify no deviations from its requirements.

\item {\bf Broader impacts}
    \item[] Question: Does the paper discuss both potential positive societal impacts and negative societal impacts of the work performed?
    \item[] Answer: \answerYes{} 
    \item[] Justification: Please refer to Section~\ref{sec:impact_statement}
    
\item {\bf Safeguards}
    \item[] Question: Does the paper describe safeguards that have been put in place for responsible release of data or models that have a high risk for misuse (e.g., pre-trained language models, image generators, or scraped datasets)?
    \item[] Answer: \answerNA{}{} 
    \item[] Justification: This paper poses no such risks.

\item {\bf Licenses for existing assets}
    \item[] Question: Are the creators or original owners of assets (e.g., code, data, models), used in the paper, properly credited and are the license and terms of use explicitly mentioned and properly respected?
    \item[] Answer: \answerYes{} 
    \item[] Justification: The information has been provided in Appendix~\ref{sec:dataset_descriptions}

\item {\bf New assets}
    \item[] Question: Are new assets introduced in the paper well documented and is the documentation provided alongside the assets?
    \item[] Answer: \answerNA{} 
    \item[] Justification: This paper does not release new assets.

\item {\bf Crowdsourcing and research with human subjects}
    \item[] Question: For crowdsourcing experiments and research with human subjects, does the paper include the full text of instructions given to participants and screenshots, if applicable, as well as details about compensation (if any)? 
    \item[] Answer: \answerNA{} 
    \item[] Justification: This paper does not involve crowdsourcing nor research with human subjects.

\item {\bf Institutional review board (IRB) approvals or equivalent for research with human subjects}
    \item[] Question: Does the paper describe potential risks incurred by study participants, whether such risks were disclosed to the subjects, and whether Institutional Review Board (IRB) approvals (or an equivalent approval/review based on the requirements of your country or institution) were obtained?
    \item[] Answer: \answerNA{} 
    \item[] Justification: The paper does not involve crowdsourcing nor research with human subjects.

\item {\bf Declaration of LLM usage}
    \item[] Question: Does the paper describe the usage of LLMs if it is an important, original, or non-standard component of the core methods in this research? Note that if the LLM is used only for writing, editing, or formatting purposes and does \emph{not} impact the core methodology, scientific rigor, or originality of the research, declaration is not required.
    \item[] Answer: \answerYes{} 
    \item[] Justification: LLMs are used as core components of FoG and their roles are described in the method and prompt-template sections. Details are provided in Section~\ref{subsec:memory}

\end{enumerate}

\label{sec:appendix-impl}

\section{Dataset Descriptions}
\label{sec:dataset_descriptions}
In this section, we describe the datasets used in our experiments: 

\begin{itemize}
    \item \textbf{WebQSP}~\cite{webqsp} 
    (\href{https://www.microsoft.com/en-us/download/details.aspx?id=52763}{available online}, 
    CC-BY 4.0 license) is a Freebase-based KBQA dataset containing natural-language questions annotated with semantic parses and answers, designed for evaluating semantic parsing and multi-hop question answering.

    \item \textbf{CWQ}~\cite{cwq} 
    (\href{https://allenai.org/data/complexwebquestions}{available online}, 
    Apache-2.0 license) is a complex question answering dataset built on Freebase, where each question is paired with annotations, requiring compositional and multi-hop reasoning.

    \item  \textbf{Freebase}~\cite{bollacker2008freebase}
    (\href{https://developers.google.com/freebase}{available online}, 
    CC-BY license) is a large-scale collaboratively constructed knowledge base originally developed by Google. 
\end{itemize}
\section{Baselines}
\label{sec:app-baselines}
We compare FoG with 19 different baseline methods, which we group into four categories: Non-LLMs, LLMs-only, RAG-based, and path-search-based methods. The details are as follows.

Baselines are grouped into 4 categories across 19 methods: Non-LLM methods, LLM-only methods, graph retrieval-based methods and LLM-guided graph agents methods. The detailed information for each baseline is as follows:

\paragraph{Non-LLM methods:}

\begin{itemize}[leftmargin=*,]
\item \textbf{NSM} \cite{he2021nsm} introduces a teacher network that learns intermediate supervision signals for multi-hop KBQA by combining forward and backward reasoning to estimate intermediate entity distributions. These signals stabilize student training and reduce spurious reasoning.

\item \textbf{SQALER} \cite{atzeni2021sqaler} decouples multi-hop logical reasoning from graph instance size and performs reasoning in a way that scales linearly with the number of relation types rather than edges. It generates candidates that can be provably refined to recover the exact solutions, enabling much better scalability.

\item \textbf{KGT5} \cite{saxena2022kgt5} reframes KG link prediction as a sequence-to-sequence generation problem, replacing triple scoring with autoregressive decoding in an encoder–decoder Transformer. This yields comparable or better performance with dramatically smaller model size than entity-embedding KGE models, and transfers well to KGQA on incomplete KGs.

\item \textbf{UniKGQA} \cite{jiang2022unikgqa} unifies subgraph retrieval and multi-hop reasoning with a shared architecture that performs PLM-based question–relation matching and propagates matching signals along directed KG edges. It further ties both stages via shared question–relation pretraining and stage-specific fine-tuning.

\end{itemize}

\textbf{LLM-only methods: }
\begin{itemize}[leftmargin=*,]

\item \textbf{Zero-shot GPT-4o} uses GPT-4o without accessing any external knowledge and answers each question directly from its parametric knowledge under a minimal task instruction. This setting serves as a cost-efficient LLM-only baseline but is prone to factual hallucination on multi-hop or entity-centric queries.

\item \textbf{Few-shot GPT-4o} augments the prompt with few in-context exemplars that demonstrate the desired answer type and output format for KBQA. By learning the task pattern from demonstrations, it improves formatting compliance and often yields more stable predictions, while still relying purely on parametric knowledge and incurring higher prompt cost.

\item \textbf{CoT GPT-4o} prompts GPT-4o to explicitly generate intermediate reasoning steps (e.g., ``let's think step by step'') before producing the final answer, aiming to better handle compositional and multi-hop questions. Although CoT can enhance logical consistency, in the LLM-only setting it may produce plausible but ungrounded reasoning and increases token usage due to verbose rationales.
\end{itemize}

\paragraph{Graph retrieval-based methods:}
\begin{itemize}[leftmargin=*,]

\item \textbf{G-Retriever} \cite{he2024gretriever} formulates graph retrieval for RAG as a Prize-Collecting Steiner Tree optimization to extract a compact, question-relevant evidence subgraph from large textual graphs. This design scales beyond the LLM context window and helps reduce hallucination.

\item \textbf{SubgraphRAG} \cite{li2025simple} proposes a lightweight MLP-based parallel triple-scoring retriever with directional structural-distance encoding for efficient subgraph retrieval. It can flexibly control retrieved subgraph size to balance effectiveness and efficiency without LLM fine-tuning.

\item \textbf{LightPROF} \cite{ao2025lightprof} proposes a “Retrieve-Embed-Reason” KGQA framework that injects KG structural + factual signals via a Transformer-based Knowledge Adapter mapped into the LLM embedding space. It trains only the adapter, enabling parameter-efficient reasoning with small open-source LLMs.

\item \textbf{GNN-RAG} \cite{mavromatis2025gnnrag} uses a GNN to retrieve answer candidates from dense KG subgraphs and extracts shortest-path reasoning chains as evidence. The paths are verbalized and fed to an LLM for final answer generation.

\end{itemize}

\paragraph{LLM-guided graph agents:}
\begin{itemize}[leftmargin=*,]

\item \textbf{ToG} \cite{thinkongraph} treats an LLM as an agent that interactively explores a KG and iteratively runs beam search to find promising reasoning paths. It is training-free and plug-and-play across LLMs/KGs/prompts while improving traceable, grounded reasoning.

\item \textbf{RoG} \cite{luo2023rog} proposes a planning–retrieval–reasoning pipeline where the LLM first generates KG-grounded relation-path plans, then retrieves matching reasoning paths from the KG as evidence. It leverages KG structure to produce more faithful and interpretable reasoning (with optional training-based distillation).

\item \textbf{PoG} \cite{chen2024pog} proposes a self-correcting adaptive planning loop that decomposes questions into sub-objectives and iteratively explores KG paths with Guidance–Memory–Reflection. This enables adaptive search breadth and correction of erroneous reasoning paths during KG-augmented reasoning.

\item \textbf{EPERM} \cite{long2025eperm} reformulates KGQA as a graphical model and introduces an evidence-path pipeline that filters and **weights** supporting reasoning paths by their importance. The LLM then reasons over these weighted evidence paths rather than treating all retrieved KG information equally.

\item \textbf{ORT} \cite{liu2025ort} proposes ontology-guided reverse thinking by extracting purpose/condition labels and building label-level reasoning paths from purposes back to conditions. These ontology-based label paths then guide KG retrieval for KGQA.

\item \textbf{FiDeLiS} \cite{sui2025fidelis} anchors LLM answers to verifiable KG reasoning steps via stepwise beam search with a deductive scoring function that validates each step and stops once the query becomes deducible. It adds a Path-RAG module to pre-select candidates per step, shrinking the search space for scalable traversal.

\item \textbf{BYOKG-RAG} \cite{mavromatis2025byokgrag} lets LLMs generate graph artifacts (entities/candidates/paths/OpenCypher queries) and uses specialized graph tools to ground and retrieve context, then iteratively refines linking and retrieval. It is designed to generalize robustly to custom (“bring-your-own”) KGs and reduce sensitivity to traversal initialization.

\item \textbf{KG-Agent} \cite{jiang-etal-2025-kg} proposes an autonomous LLM agent that iteratively selects tools and updates a knowledge memory to execute multi-hop KG reasoning. It represents reasoning as programs and fine-tunes a small LLM using a synthesized code-instruction dataset.
\end{itemize}

\begin{figure*}[htbp]
    \centering
    \includegraphics[width=1\textwidth]{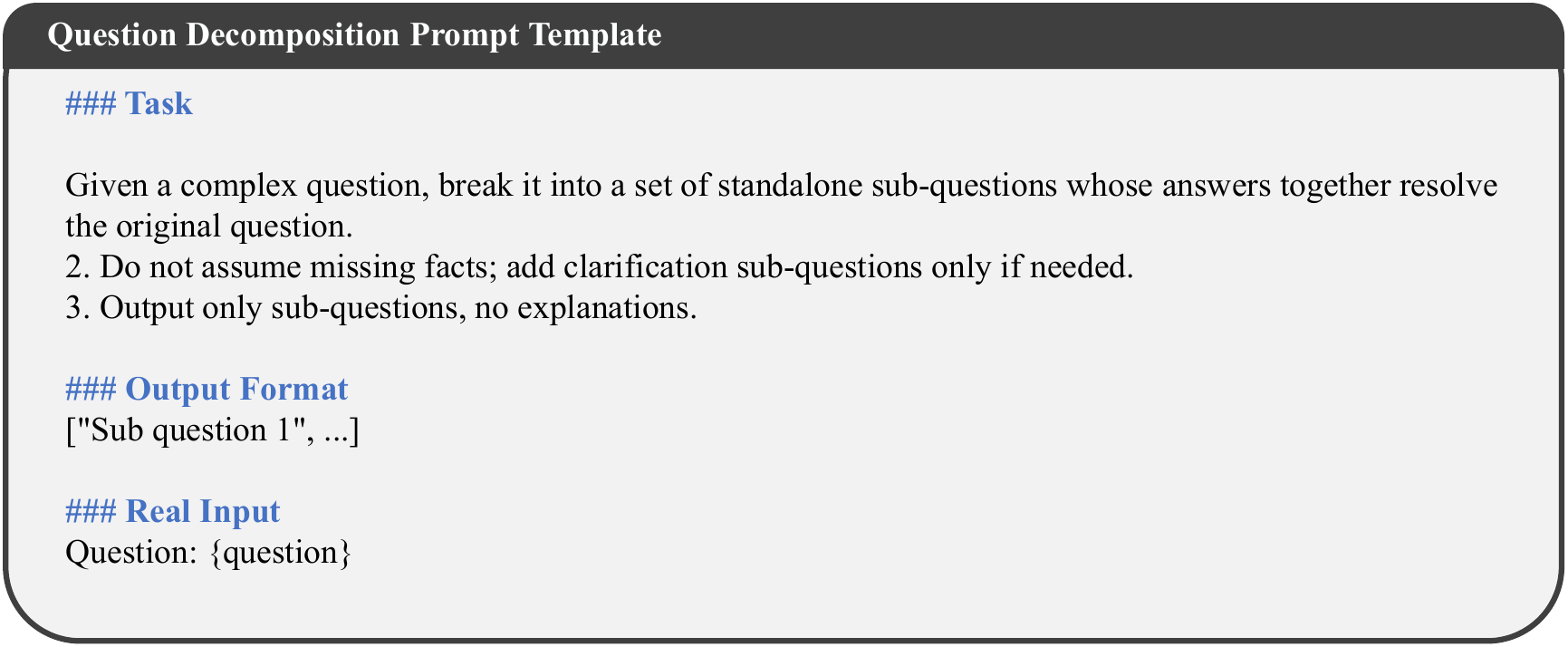}
    \caption{The prompt template of question decomposition prompt.}
    \label{prompt:subquestion-decomposion}
\end{figure*}

  \begin{figure*}[htbp]
    \centering
    \includegraphics[width=1\textwidth]{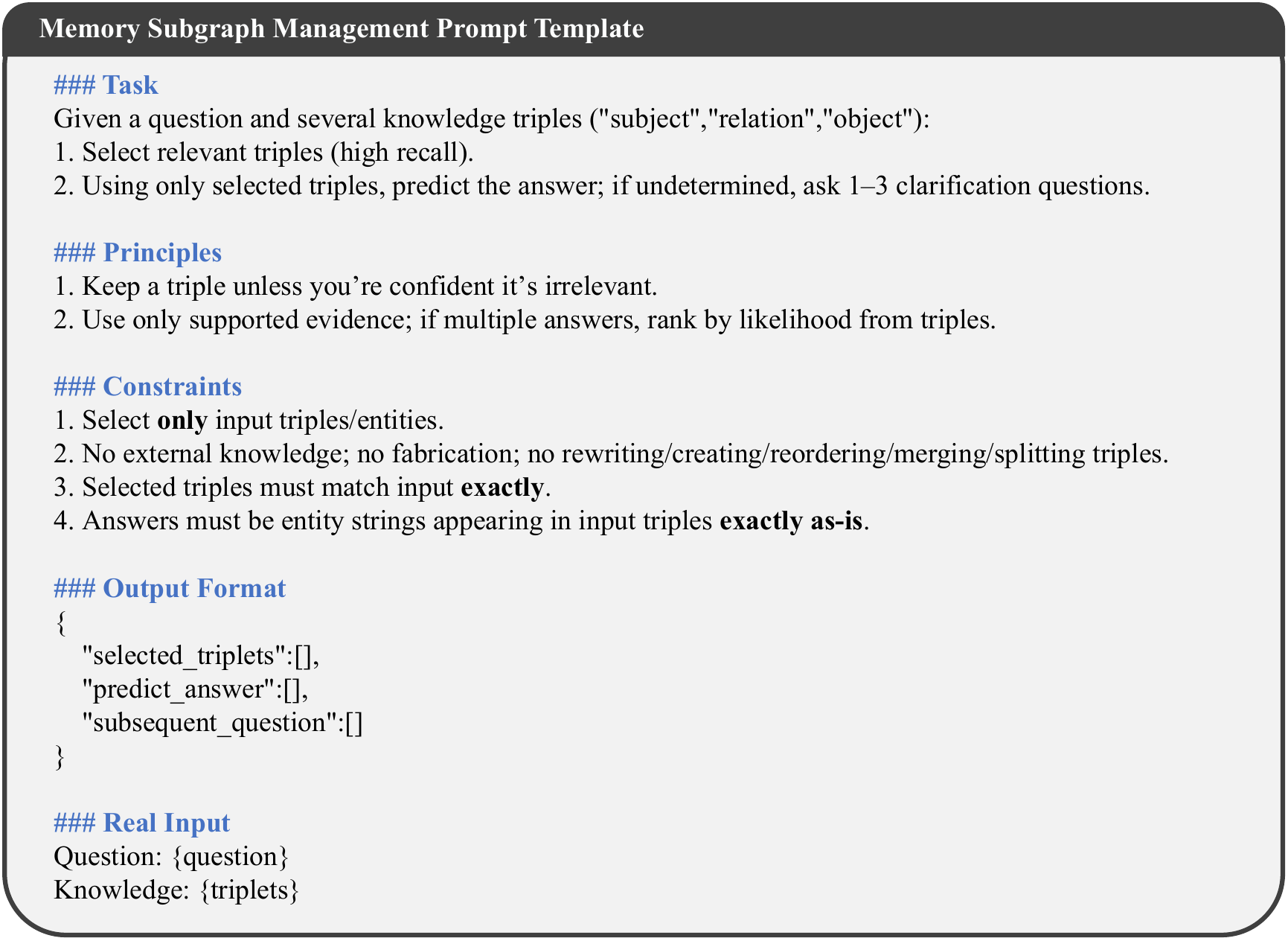}
    \caption{The prompt template of memory subgraph management.}
    \label{prompt:mem}
\end{figure*}

\section{LLM Prompt Template}
\label{sec:prompt-template}
In this section, we illustrate all the templates and prompts used in the experiments.

\paragraph{Question decomposition prompt template} We use an instruction prompt to decompose the original question into a list of standalone sub-questions whose answers jointly resolve the original question. The prompt prohibits assuming missing facts and allows clarification sub-questions only when necessary. The output is restricted to a JSON-style list ["Sub question 1", ...], with \{question\} replaced by the input question. The prompt template is provided in Figure~\ref{prompt:subquestion-decomposion}.

\paragraph{Memory subgraph management prompt template} This prompt takes the input question and a set of KG triples in the form ((\text{subject}, \text{relation}, \text{object})). It instructs the LLM to select a high-recall subset of relevant triples, predict the answer using only the selected triples (or leave it undetermined), and ask 1–3 clarification sub-questions when the answer cannot be concluded. The prompt enforces that selected triples and answer entities must be copied exactly from the inputs, without using external knowledge or modifying any triples. The output is a JSON object with fields "selected\_triplets", "predict\_answer", and "subsequent\_question".
The prompt template is provided in Figure~\ref{prompt:mem}.

\label{sec:appendix-llm}

\begin{figure*}[htbp]
    \centering
    \includegraphics[width=1\textwidth]{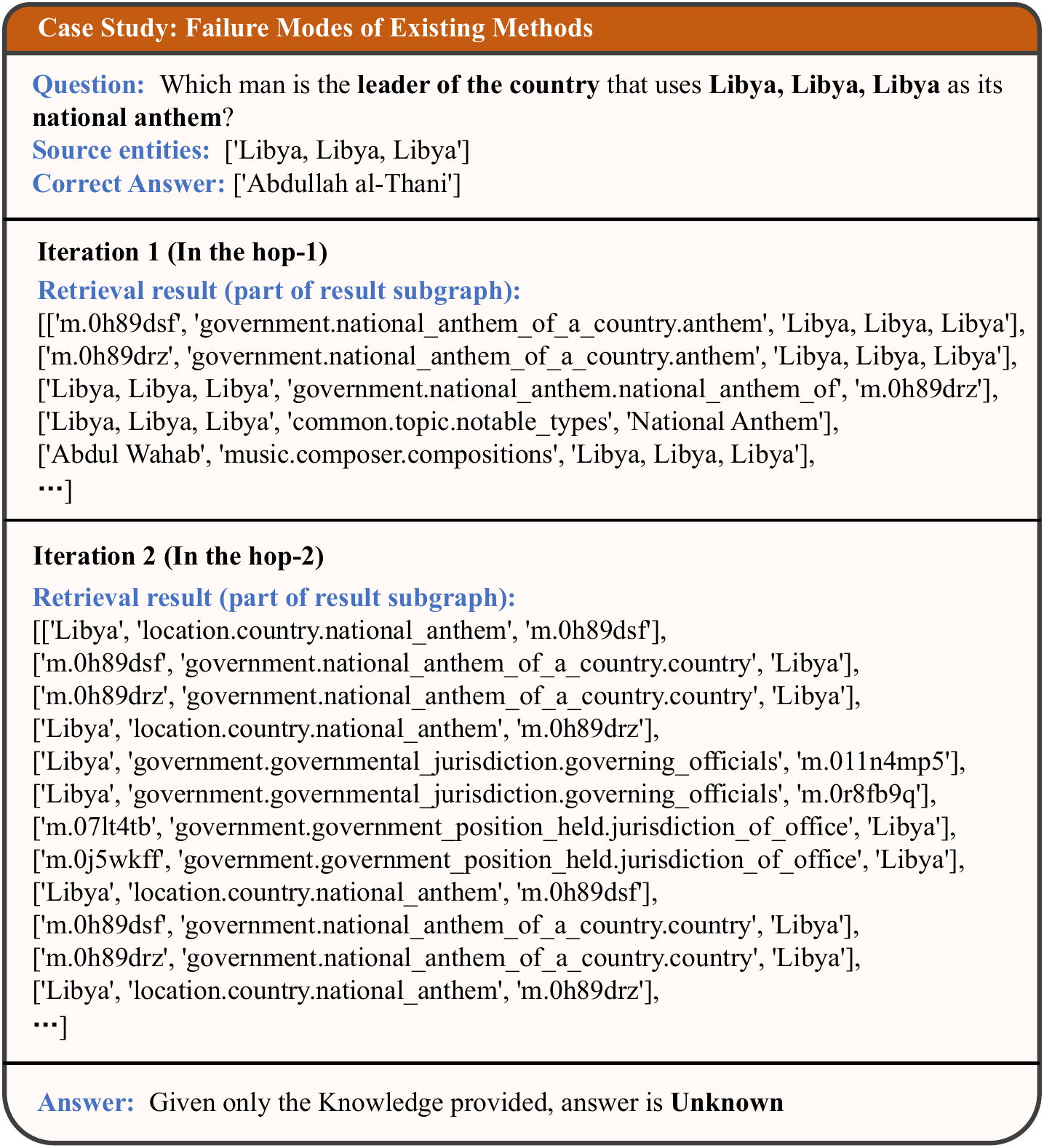}
    \caption{Case study 1: failure modes of existing methods.}
    \label{case:1}
\end{figure*}

\begin{figure*}[!htbp]
    \centering
    \includegraphics[width=1\textwidth]{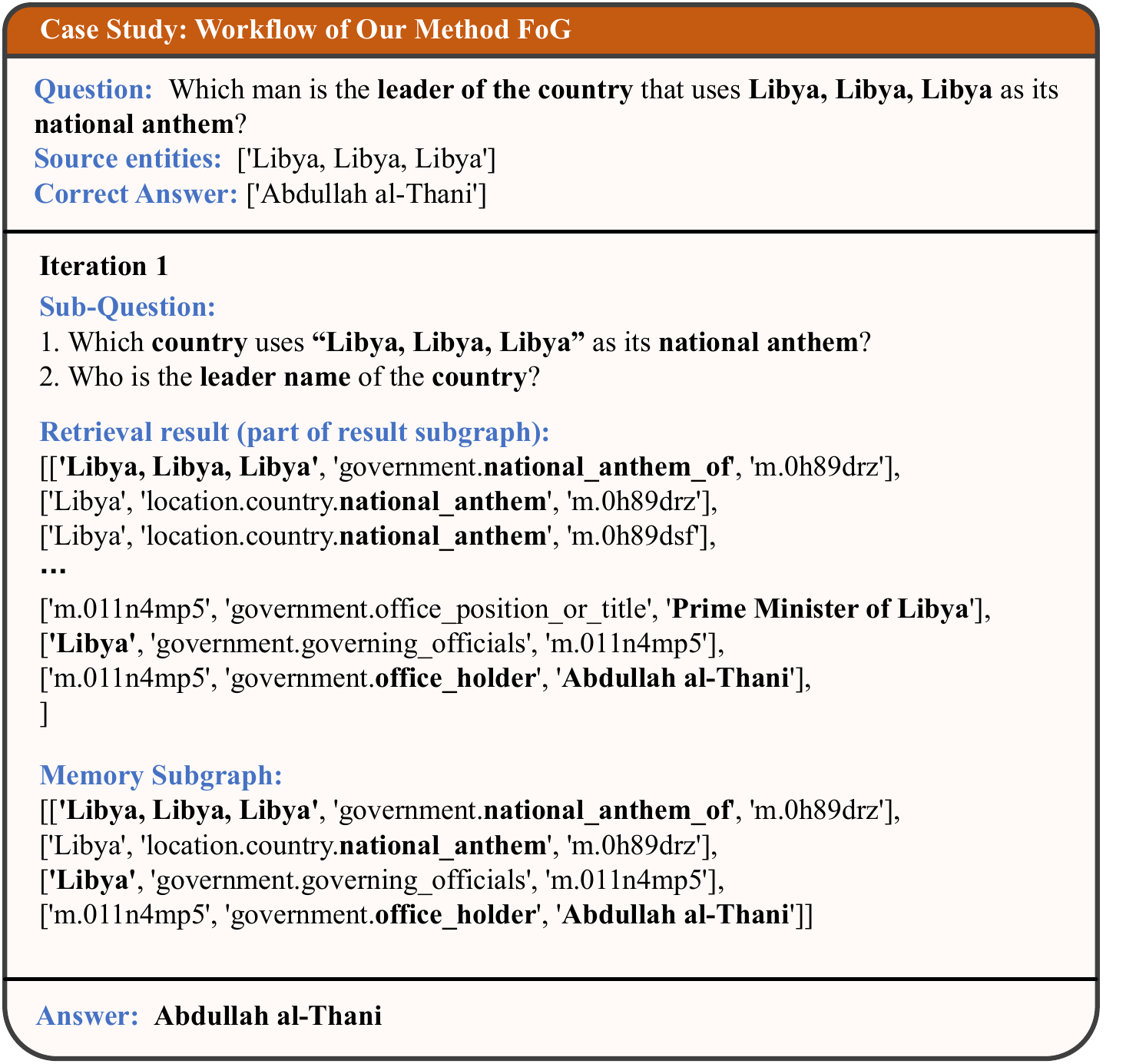}
    \caption{Case study 2: workflow of our method FoG.}
    \label{case:2}
\end{figure*}
\section{Case study}
\label{app:case-study}
In this case study, we contrast standard hop-by-hop KG traversal with FoG. As shown in Figure \ref{case:1}, greedy/beam-style per-hop pruning is prone to a myopic failure mode on large-scale KGs: in the early search stages, the model has little or no access to distant, answer-critical entities, so many outgoing relations appear similarly relevant under local matching. Consequently, truly answer-relevant relations may not look sufficiently salient and can be mistakenly pruned by top-$k$/beam selection. Once these critical edges are removed, subsequent exploration cannot recover the missing evidence chain, causing the search to drift into irrelevant neighborhoods and ultimately fail. 

In contrast, Figure \ref{case:2} illustrates how FoG mitigates this issue. Starting from the extracted source entities, FoG iteratively expands a subgraph while discarding question-irrelevant triples in each iteration, yielding an expanded subgraph with high recall and low noise. More importantly, FoG introduces a far-to-near message feedback mechanism that propagates strong question-relevant signals discovered at distant entities back toward nodes closer to the sources, and uses these lookahead cues to re-score nearby triples, thereby preventing incorrect early-stage pruning. Finally, the LLM incorporates the relevance-refined graph structure into its memory and generates follow-up sub-questions to guide the next retrieval iteration, enabling the search to reliably converge along the correct evidence chain to the answer.

\section{Training Details}
\label{sec:appendix-training}

This subsection supplements \S\ref{subsec:training} by providing additional details on how we train the foresight triple scorer with answer-supporting evidence and local noise sampling.

\paragraph{Training Evidence}
\label{sec:appendix-evidence-extraction}

For each training question $q_{\mathrm{tr}}$, let $\mathcal{G}^{+}(q_{\mathrm{tr}})$ denote the set of triples on its answer-supporting reasoning paths in the knowledge graph. These triples serve as positive evidence for training the foresight triple scorer. For readability, we omit the explicit dependence on $q_{\mathrm{tr}}$ in the remainder of this subsection and use $\mathcal{T}^{\mathrm{pos}}$ to denote the positive evidence triples for the current training instance:
\begin{equation}
\mathcal{T}^{\mathrm{pos}} = \mathcal{G}^{+}(q_{\mathrm{tr}}).
\end{equation}

The foresight scorer is trained with triple-level supervision. Triples in $\mathcal{T}^{\mathrm{pos}}$ are labeled as positive evidence, while locally sampled neighboring triples outside $\mathcal{T}^{\mathrm{pos}}$ are used as negative candidates.

\paragraph{Local Training Region and Noise Sampling}
\label{sec:appendix-local-training}

To train the foresight scorer in a local context consistent with inference-time retrieval, we construct a per-question local training region by sampling noisy neighboring triples around the positive evidence. We initialize the seed entity set as the entities touched by the positive evidence triples, optionally unioned with an extra anchored set $\mathcal{E}_0$ obtained from preprocessing:
\begin{equation}
\mathcal{E}^{\mathrm{seed}}
=
\{v \mid \exists\, t \in \mathcal{T}^{\mathrm{pos}},\ v \in t\}
\cup
\mathcal{E}_0 .
\end{equation}

For each seed entity $v\in\mathcal{E}^{\mathrm{seed}}$, we consider its one-hop neighboring triples $\mathcal{N}(v)$ and sample noisy triples around $v$ proportionally to how often $v$ participates in the positive evidence. Let
\begin{equation}
\deg_{\mathrm{pos}}(v)
=
\bigl|\{t\in\mathcal{T}^{\mathrm{pos}}:\ v\in t\}\bigr|.
\end{equation}

We uniformly sample $\kappa\cdot \deg_{\mathrm{pos}}(v)$ triples from $\mathcal{N}(v)\setminus \mathcal{T}^{\mathrm{pos}}$ as noisy candidates. The sampled triples are deduplicated across all seed entities to form the negative set $\mathcal{T}^{\mathrm{neg}}$:
\begin{equation}
\mathcal{T}^{\mathrm{neg}}
=
\bigcup_{v\in\mathcal{E}^{\mathrm{seed}}}
\mathrm{Sample}
\left(
\mathcal{N}(v)\setminus \mathcal{T}^{\mathrm{pos}},
\kappa\cdot \deg_{\mathrm{pos}}(v)
\right).
\end{equation}

We set $\kappa=10$ in all experiments. To reduce uninformative noise, we optionally filter out triples triggered by extremely high-degree generic relations before sampling.

The training triples for each instance are then defined as:
\begin{equation}
\mathcal{T}^{\mathrm{train}}
=
\mathcal{T}^{\mathrm{pos}}
\cup
\mathcal{T}^{\mathrm{neg}}.
\end{equation}

Each triple $t\in\mathcal{T}^{\mathrm{train}}$ is assigned an edge-level binary label:
\begin{equation}
y_t
=
\mathbb{I}\!\left[t\in\mathcal{T}^{\mathrm{pos}}\right].
\end{equation}

Thus, triples on answer-supporting reasoning paths are treated as positive examples, while sampled neighboring triples outside these paths are treated as negative examples.

\paragraph{Objective and Optimization}
\label{sec:appendix-optim}

We obtain a foresight score $s_t$ for each triple $t$ following \S\ref{subsec:foresight-msg}, and interpret $\sigma(s_t)$ as the predicted probability that $t$ belongs to an answer-supporting reasoning path. For each training question $q_{\mathrm{tr}}$, we compute a triple-level binary cross-entropy loss over its local training set $\mathcal{T}^{\mathrm{train}}$:
\begin{equation}
\begin{aligned}
\mathcal{L}_{\mathrm{triple}}(q_{\mathrm{tr}})
=
\sum_{t\in\mathcal{T}^{\mathrm{train}}}
\Big(
&- y_t \log \sigma(s_t) \\
&- (1-y_t)\log\bigl(1-\sigma(s_t)\bigr)
\Big).
\end{aligned}
\end{equation}

For a mini-batch of training questions $\mathcal{B}$, the final training objective is:
\begin{equation}
\mathcal{L}_{\mathrm{tr}}
=
\frac{1}{|\mathcal{B}|}
\sum_{q_{\mathrm{tr}}\in\mathcal{B}}
\mathcal{L}_{\mathrm{triple}}(q_{\mathrm{tr}}).
\end{equation}

We optimize model parameters with mini-batch AdamW by minimizing $\mathcal{L}_{\mathrm{tr}}$, while keeping the underlying embedding encoder frozen throughout training. We use AdamW with learning rate $1\times 10^{-5}$, $\beta_1=0.9$, $\beta_2=0.999$, and weight decay $1\times 10^{-3}$, without learning rate warmup or decay. We apply dropout with rate $0.1$ to the foresight feedback propagation module and the foresight triple scorer.

We train for $5$ epochs on each dataset. All experiments are conducted on a single NVIDIA GeForce RTX 5090 GPU.

\paragraph{Training and Inference Separation}

The above triple-level labels are used only for supervised training of the foresight scorer. During inference, FoG follows the retrieval procedure described in Section~\ref{sec:method}: it starts from the question and the extracted source entities, expands the knowledge graph with the learned foresight scorer, applies far-to-near feedback propagation, and maintains a compact memory subgraph for answer generation.

\section{Impact Statement}
\label{sec:impact_statement}
Foresight-over-Graph (FoG) introduces a foresight-aware retrieval framework for knowledge-graph-grounded question answering, enabling large language model (LLM)-based agents to explore complex knowledge graphs beyond local horizons, preserve answer-critical evidence, and reason over multi-hop factual structures with improved reliability and efficiency. The broader impacts of this work include advancing trustworthy and interpretable knowledge-intensive AI systems, with potential applications in scientific discovery, professional decision support, enterprise knowledge management, and collaborative AI assistants that require accurate reasoning over structured, updatable knowledge sources. However, if the underlying knowledge graph contains biased, incomplete, outdated, or adversarially injected facts, FoG may propagate and reinforce misleading evidence through its retrieval and memory mechanisms. We therefore urge responsible deployment of this framework with appropriate safeguards, including continual knowledge validation, provenance tracking, robustness checks against noisy or malicious graph entries, and alignment with human oversight in high-stakes applications.

\end{document}